%% file: main-8463-Liu.tex
\documentclass[11pt,a4paper]{article}
\usepackage{times,latexsym}
\usepackage{url}
\usepackage[T1]{fontenc}

\usepackage[acceptedWithA]{tacl2021v1}
\usepackage[]{tacl2021v1}
\usepackage{times}
\usepackage{latexsym}
\usepackage{inconsolata}
\usepackage[utf8]{inputenc}
\usepackage{microtype}
\usepackage{inconsolata}
\usepackage{graphicx}
\usepackage{tcolorbox}
\usepackage{inconsolata}
\usepackage{microtype}
\usepackage{enumitem}
\usepackage{geometry}
\usepackage{amssymb}
\usepackage{mathrsfs}
\usepackage{amsmath}
\usepackage{hhline}
\usepackage{amsfonts}
\usepackage{multirow}
\usepackage{tabularx}
\usepackage{array}
\usepackage{natbib}
\usepackage{booktabs}
\usepackage{verbdef}
\usepackage{colortbl}
\usepackage{color}
\usepackage{subcaption}
\usepackage{threeparttable}
\usepackage{float}
\usepackage{caption}
\usepackage{arydshln}
\usepackage{inconsolata}
\usepackage{rotating}
\usepackage{xcolor}
\usepackage{hyperref}
\usepackage{placeins}
\usepackage{makecell}
\usepackage{afterpage}
\usepackage{blindtext}
\usepackage{threeparttablex}
\usepackage[framemethod=TikZ]{mdframed}
\usepackage{hhline}
\definecolor{mygray}{gray}{0.9}
\definecolor{mypink}{rgb}{0.99,0.91,0.95}
\definecolor{Coral}{RGB}{245, 218, 210}
\definecolor{mycyan}{cmyk}{0.3,0,0,0}
\definecolor{Celadon}{RGB}{172, 225, 175}
\definecolor{Peach}{RGB}{255, 229, 180}
\definecolor{orange}{HTML}{FFF8E3}
\definecolor{green}{HTML}{F3FFF0}
\definecolor{blue}{HTML}{EAF8FF}
\newcolumntype{P}[1]{>{\raggedright\arraybackslash}p{#1}}

\usepackage{xspace,mfirstuc,tabulary}

\newif\iftaclinstructions
\taclinstructionsfalse 
\iftaclinstructions
\renewcommand{\confidential}{}
\renewcommand{\anonsubtext}{(No author info supplied here, for consistency with
TACL-submission anonymization requirements)}
\newcommand{\instr}
\fi

\iftaclpubformat 

\else

\fi

\title{Explanatory Summarization with Discourse-Driven Planning}

\author{
 \textbf{Dongqi Liu\textsuperscript{$\Omega$}\thanks{\; This work was conducted during the research visit at the University of Edinburgh.}}\ ,
 \textbf{Xi Yu\textsuperscript{$\Omega$}},
 \textbf{Vera Demberg\textsuperscript{$\Omega$}},
 \textbf{Mirella Lapata\textsuperscript{$\Theta$}} \\
 \textsuperscript{$\Omega$}Saarland University,
 \textsuperscript{$\Theta$}University of Edinburgh \\
 {
   \shortstack[c]{
     \textsuperscript{$\Omega$}\texttt{\{dongqi,xiyu,vera\}@lst.uni-saarland.de} \\
     \textsuperscript{$\Theta$}\texttt{mlap@inf.ed.ac.uk}
   }
 }
}

\date{}

\begin{document}
\maketitle
\begin{abstract}
Lay summaries for scientific documents typically include explanations to help readers grasp sophisticated concepts or arguments. However, current automatic summarization methods do not explicitly model explanations, which makes it difficult to align the proportion of explanatory content with human-written summaries. In this paper, we present a plan-based approach that leverages discourse frameworks to organize summary generation and guide explanatory sentences by prompting responses to the plan. Specifically, we propose two discourse-driven planning strategies, where the plan is conditioned as part of the input or part of the output prefix, respectively. Empirical experiments on three lay summarization datasets show that our approach outperforms existing state-of-the-art methods in terms of summary quality, and it enhances model robustness, controllability, and mitigates hallucination. The project information is available at \url{https://dongqi.me/projects/ExpSum}.
\end{abstract}

\input{sections/introduction}

\input{sections/related_work}
\input{sections/approach}
\input{sections/experiments}
\input{sections/analysis}
\input{sections/conclusion}

\bibliography{custom}
\bibliographystyle{acl_natbib}
\appendix

\input{sections/appendix}
\end{document}

%% file: sections/introduction.tex
\section{Introduction}

\begin{figure}[t]
  \centering
  \includegraphics[width=0.5\textwidth]{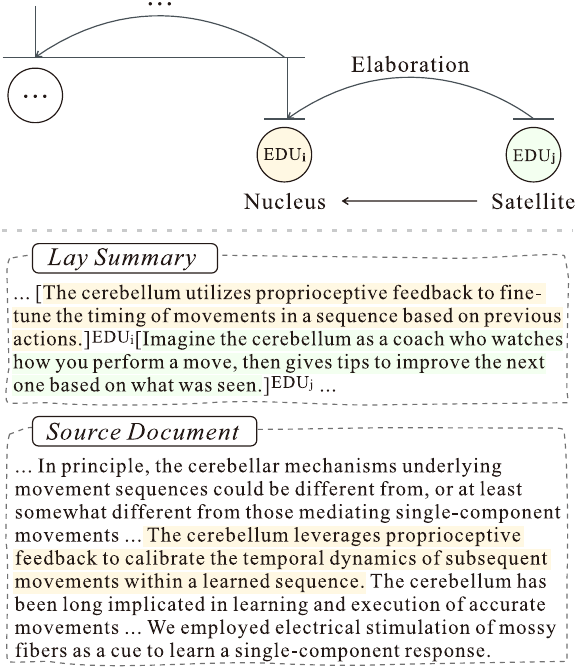}
  \caption{An excerpt of a source document paired with its summary (bottom); the explanatory sentence and its target are highlighted in \colorbox{green}{green} and \colorbox{orange}{orange}, respectively. The RST tree (top) for the text corresponding to the two sentences shows they are linked by the rhetorical relation \emph{Elaboration}.}
  \label{fig:example}
\end{figure}

In the domain of lay summarization for scientific documents, the inclusion of explanatory content is beneficial for improving readability and accessibility, particularly when dealing with difficult concepts or complicated statements \cite{srikanth-li-2021-elaborative, august-etal-2022-generating, luo-etal-2022-readability, goldsack-etal-2022-making}. As an example, consider the target sentence being explained (highlighted in \colorbox{orange}{orange}) within the lay summary in Figure~\ref{fig:example}. It describes the role of the cerebellum (a similar sentence can be found in the source document, also highlighted in \colorbox{orange}{orange}) and is accompanied by an explanation highlighted in \colorbox{green}{green}, which compares the cerebellum to a coach. This analogy, which illustrates how the cerebellum adjusts the timing of actions based on previous movements, makes the preceding technical description easier to understand. Using simpler vocabulary or shorter sentences can make the summaries easier to read, but could also lead to misinterpretations \cite{pu-demberg-2023-chatgpt, han-etal-2023-bridging, wu-etal-2023-elaborative, hewett-etal-2024-elaborative}. Thus, explanations are often used to avoid this, balancing accessibility with accuracy and cognitive load \cite{oksa2010expertise}.

Empirical analysis of lay summaries supports the observation that explanations are commonplace. Specifically, by using a discourse parser \cite{liu-etal-2020-multilingual-neural, liu-etal-2021-dmrst} to identify explanations in the eLife and PLOS (lay summarization) datasets \cite{goldsack-etal-2022-making}, we find that explanations account for approximately~19.02\% and 18.19\% of summary sentences, respectively. This proportion of explanations in lay summaries is three times higher than in expert summaries (6.28\% and 5.06\%) and nearly four times greater than in the original academic papers (5.16\% and 4.76\%). In other words, explanations are significantly more common than other discourse relations like \emph{Condition} and \emph{Purpose}. This underscores their importance in structuring lay summaries and enhancing text accessibility.

Most lay summarization models follow an end-to-end approach \cite{goldsack-etal-2022-making, pu-etal-2022-two, goldsack-etal-2023-enhancing, pu2024scinews} without \emph{explicitly} accounting for explanations. As a result, explanations are often underrepresented in generated summaries, which in turn may be the reason why generated summaries lack the clarity and readability found in human-produced ones \cite{guo2021automated, goldsack-etal-2023-enhancing, tang-etal-2023-improving, pu-demberg-2023-chatgpt, zhang-etal-2024-atlas, wang2024science}. In this paper, we develop neural models that are capable of generating \textit{lay summaries} with controlled explanatory content. We achieve this by \emph{planning} the content of the summary, thus \emph{directly} steering the model towards generating explanations. Plan-based models are a suitable type of approach for lay summarization, as they are less prone to hallucinations, offer greater controllability \citep{Puduppully_Dong_Lapata_2019,narayan-etal-2021-planning,moryossef-etal-2023-open,narayan-etal-2023-conditional,huot-etal-2023-text,huot-etal-2024-mplan, liu2025talkaboutvideototextsummarization}, and can be flexibly adapted to different generation tasks depending on how plans are formulated. 

Our method conceptualizes plans as a series of questions that trigger explanatory content, with their strategic placement learned from training data. This placement is not arbitrary; rather, it reflects where human writers tend to insert explanations to clarify complex claims or findings. Importantly, our approach not only models where to explain, but also what is being explained and how—e.g., through analogy, background, or causal reasoning. In addition, evaluating such explanations remains challenging, as automatic consistency detection metrics often misclassify useful additions as hallucinations \cite{cao-etal-2022-hallucinated}. To address this, we incorporate external knowledge verification, but acknowledge that finer-grained human or LLM-based judgments are still needed for nuanced evaluation.

A common challenge with plan-based models is the scarcity of datasets with plan annotations. As a result, plans are often reverse-engineered from existing reference summaries through automatic means, e.g.,~by identifying entities \cite{narayan-etal-2021-planning,huot-etal-2024-mplan} or generating questions and their answers \cite{narayan-etal-2023-conditional}. Our approach draws inspiration from Rhetorical Structure Theory (RST; \citealt{mann1987rhetorical}). RST represents texts as trees (see Figure~\ref{fig:example}), where the leaves correspond to Elementary Discourse Units (EDUs) and the nodes specify how these and larger units are connected through rhetorical relations (e.g.,~\emph{Elaboration}). Discourse units are further characterized in terms of their text importance: \emph{nuclei} represent central segments, whereas \emph{satellites} denote peripheral ones. As depicted in the upper part of Figure~\ref{fig:example}, EDU$_j$ acts as a satellite to~EDU$_i$, supplying an explanation. 

\begin{figure*}[t]
  \centering
  \includegraphics[width=1\textwidth]{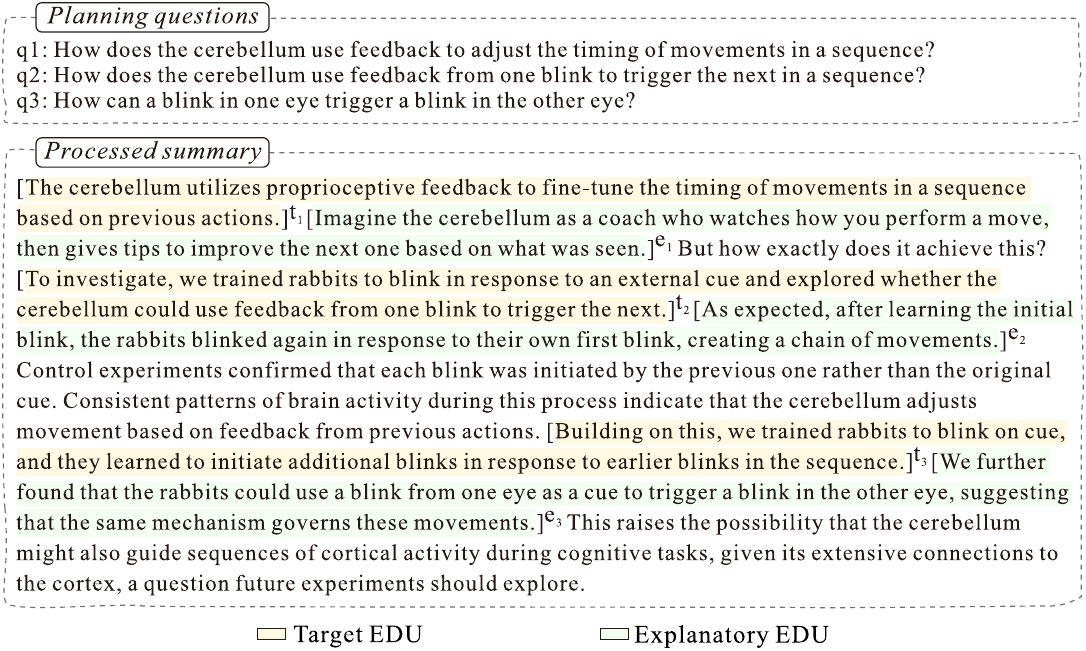}
  \caption{We use \texttt{DMRST} \cite{liu-etal-2021-dmrst} to extract \colorbox{green}{explanatory ($e$) EDUs} and their \colorbox{orange}{target ($t$) EDUs} from reference summaries, and then feed this data into \texttt{GPT-4o} to generate plans ($b$).}
  \label{fig:preparatation_phase}
\end{figure*}

We automatically generate plans from reference summaries annotated with RST trees \cite{liu-etal-2020-multilingual-neural, liu-etal-2021-dmrst}. We hypothesize that explanatory EDUs answer latent questions \cite{beaver2017questions}, which we verbalize. Our plans are designed to pose these latent questions, but we do not state the answers explicitly (see Figure~\ref{fig:preparatation_phase}). We develop two model variants that differ in how they integrate planning in the summarization process, with plans either included as part of the source document or as part of the target summary. Empirical results using \texttt{Mistral} \cite{jiang2023mistral7b} as backbone confirm that our discourse-driven planning approach significantly improves the quality of lay summaries but also has the potential to control the generation of explanatory content to a certain extent. Our key contributions are as follows:

\begin{itemize}[leftmargin=8pt,itemsep=1pt,topsep=1pt,parsep=1pt]
    \item We introduce the task of \textit{explanatory summarization}, which aims to generate lay summaries with controlled explanations.
    \item We propose two discourse-driven models that generate lay summaries without requiring manual annotations. Empirical results show improvements in summary quality and better alignment with human performance.
    \item Both qualitative and quantitative analyses indicate that our models mitigate content hallucinations and enhance factual accuracy. However, a gap still exists compared to human performance.
\end{itemize}

%% file: sections/related_work.tex
\section{Related Work}

\paragraph{Lay Summarization} Lay Summarization involves transforming complicated scientific texts into accessible summaries for non-experts, thereby enhancing public understanding of scientific research. Existing investigations have tackled several challenges, such as creating datasets and developing task-specific models. \citet{chandrasekaran-etal-2020-overview-insights} pioneered the task of scientific lay summarization, while follow-up work focused on the development of larger datasets, including CDSR \cite{guo2021automated}, eLife, PLOS \cite{goldsack-etal-2022-making}, and SciNews \cite{pu2024scinews}. Methodological advancements include integrating domain-specific knowledge graphs with encoder-decoder models \cite{goldsack-etal-2023-enhancing} and developing methods like ATLAS \cite{Zhang2024ATLASIL}, which facilitate the generation of summaries with varying degrees of accessibility for lay audiences. However, none of the existing approaches emphasize discourse-driven mechanisms for modulating summary content, nor do they focus on evaluating the quality or proportion of explanatory content. 

\paragraph{Interactive Summarization} Interactive Summarization refers to the process of enabling users to engage with and influence the summarization process through interactive interfaces \cite{costa2018automatic, guo2021conditional, august2023paper, kim2019designing, 10.1145/3654777.3676397}. For example, \citet{august2023paper} introduced an interface that provides localized plain-language key points, definitions, and question-based navigation, allowing users to explore medical research more effectively. Similarly, \citet{10.1145/3654777.3676397} proposed a design for recursively expandable abstracts, where users can click on keywords or phrases to access system-generated text and citation links. While interactive summarization is not the focus of our work, it is worth pointing out that users could choose to modify the plan and influence the generation process.

\paragraph{RST in Summarization}
Rhetorical Structure Theory (RST) is a discourse framework that analyzes the relations between text units, termed Elementary Discourse Units (EDUs), and their roles within the document \cite{marcu1997discourse, marcu1999discourse, marcu2000theory}. RST categorizes connections among EDUs into different rhetorical or coherence relations, such as \emph{Elaboration}, \emph{Contrast}, and \emph{Causality}. Within this framework, EDUs are arranged into a hierarchical tree structure, where nucleus EDUs are supported or elaborated by satellite EDUs. Early research \cite{marcu1997discourse, marcu1999discourse,louis-etal-2010-discourse} has found that human-written summaries are often derived from nucleus EDUs, affirming RST's usefulness for summarization tasks. Several studies have explored ways to inject discourse structure into summarization models and demonstrated improved performance compared to models without such discourse injection \cite{xu-etal-2020-discourse, dong-etal-2021-discourse, chen-yang-2021-structure, pu-etal-2023-incorporating, pu-demberg-2024-rstlora}. Our work applies RST to discern explanatory EDUs, which elucidate or support propositions, and their related target (explained) EDUs within reference summaries.

\paragraph{Planning in Summarization}
Text summarization with planning involves organizing content through structured representations prior to generating summaries. Existing work has primarily focused on \emph{phrase-level} planning, which has been shown to improve summary quality and reduce hallucination \cite{narayan-etal-2021-planning,narayan-etal-2023-conditional}. Plans have previously taken the form of named entities \cite{narayan-etal-2021-planning, liu-chen-2021-controllable, huot-etal-2024-mplan}, keyword prompts \cite{creo2023prompting}, and question-answer pairs, referred to as blueprint \cite{narayan-etal-2023-conditional}. Our work introduces \emph{EDU-level} planning, placing emphasis on larger semantic units. Our plans take the form of questions (without answers), triggered by explanations identified through RST. This definition is more flexible (compared to question-answer pairs), as we do not require answers to be incorporated and are not limited to a specific style of answers or questions. 

\paragraph{Question Under Discussion} Question Under Discussion (QUD) conceptualizes discourse as a dynamic interaction of continuously posed and answered questions \cite{roberts2012information, benz2017questions, de-kuthy-etal-2020-towards, ko-etal-2023-discourse}. It can be seen as a question generation strategy, where each sentence serves as a response to an implicit question arising from its preceding context, thereby delineating the text's intent and structure \cite{beaver2017questions, wu-etal-2023-qudeval, wu-etal-2023-elaborative}. Although QUD has been explored in discourse analysis \cite{ko-etal-2023-discourse, wu-etal-2023-qudeval} and to interpret complex linguistic phenomena like presupposition and information structure \cite{beaver2017questions}, its application in text generation and summarization has received limited attention. \citet{wu-etal-2023-elaborative} leveraged manual annotation to recover implicit questions to guide sentence-level simplification. Our work draws inspiration from QUD in formulating plans as a series of questions. We work with more complex documents and rely on large language models, bypassing issues of scalability arising from manual annotations. 

%% file: sections/approach.tex
\section{Explanatory Summarization}

\subsection{Task Definition}
We formalize \emph{explanatory summarization} as follows: Let \( D = \{(x_1, s_1), (x_2, s_2), \ldots, (x_n, s_n)\} \) denote a dataset where each tuple \((x_i, s_i)\) contains a document \( x_i \) and its corresponding lay summary~\( s_i \). We further assume $s_i$ contains some explanatory elements~\( e_i \) such as examples, contextual explanations, or background information. Our objective is to devise a summarization model that can model the generation of this explanatory content.

\subsection{Model Overview} 
Inspired by the blueprint approach \cite{narayan-etal-2023-conditional}, we propose a discourse-driven planning method to achieve the above goal. This involves transforming the original dataset \( D = \{(x_1, s_1), (x_2, s_2), \ldots, (x_n, s_n)\} \) into \( D' = \{(x_1, b_1, s_1), (x_2, b_2, s_2), \ldots, (x_n, b_n, s_n)\} \) by introducing planning component \( b \). We then train a summarization model to learn the conditional probability distribution \( P(S|X, B) \) or \( P(B, S|X) \), where \( S = \{s_1, s_2, \ldots, s_n\} \), \( X = \{x_1, x_2, \ldots, x_n\} \), and \( B = \{b_1, b_2, \ldots, b_n\} \). Each \( b_i \) in \( B \) consists of a series of ordered plan questions \( (q_1, q_2, \ldots, q_n) \), which do not initially exist in the original dataset \( D \).

Note that \( P(S|X, B) \) assumes plan \( b \) is treated as part of the input sequence, whereas \( P(B, S|X) \) treats~\( b \) as a prefix to summary generation. The key challenge in training either model is that plan \( b \) is latent. We thus automatically augment the original training pairs \( (x, s) \) into triplets \( (x, b, s) \), with  \( b \) acting as a conditional or control signal, offering an overarching plan for summary generation while also guiding the creation of explanatory sentences by prompting responses to the plan. 

\subsection{Explanatory Content Extraction}
\label{sec:content_extraction}
To obtain triplets $(x,b,s)$, we apply \texttt{DMRST} \cite{liu-etal-2020-multilingual-neural, liu-etal-2021-dmrst}, an RST-based parser, to automatically identify and extract all EDUs functioning as explanations in reference summary~$s$ (see Figure~\ref{fig:preparatation_phase}). Out of 18 rhetorical categories, we select four that are relevant to explanatory sentences: \textit{Background}, \textit{Elaboration}, \textit{Explanation}, and \textit{Comparison} (Table~\ref{tab:RST_explanatory_label} in Appendix~\ref{app:rst_labels} has the full list). Given the directed nature of RST relations, we can identify discourse units that serve as explanations and the content they explain (see Figure~\ref{fig:example}). Note that the extracted EDU pairs act only as a silver standard proxy for the subsequent training phase of our models. While our approach is parser-agnostic, we also conduct experiments with alternative parsers (see Section \ref{parser} for details).

\subsection{Explanatory Plan Generation} 
\label{sec:plan_generation}
We propose that explanatory sentences can serve as responses to (implicit) questions raised by preceding target sentences, aligning with the theoretical framework of  QUD. We employ \texttt{GPT-4o} \cite{achiam2023gpt} to generate these questions from target sentences (attested in reference summary) and the preceding context, as illustrated in Figure~\ref{fig:preparatation_phase}. For example, the plan question~q$_{2}$ is generated based on the target sentence~t$_2$ and the sentences that precede it. We do not filter the generated questions, but we create the plan by following the order in which explanations appear in the reference. The prompt for generating plan questions can be found in Appendix Figure~\ref{GPT-4o/mistral_Generate_Question_from_Answer}.

\subsection{Explanatory Summarization Models}
We propose two model variants. The first variant (\texttt{Plan-Output}) concatenates all plan questions \(b\) with the summary \(s\) to form a sequence \([b;s]\) for each document-summary pair. The training objective for this model is to generate both the plan and the summary based on the input document~\(x\). 

The second variant (\texttt{Plan-Input}) is not an end-to-end model. Instead, it employs a Plan Generation (\texttt{PG}) module and a Summary Generation (\texttt{SG}) module, which are trained separately. The \texttt{PG} module learns to generate plan~\(b\) from the input document~\(x\), and is trained on \((x,b)\) pairs, where $x$~is the input and $b$~is the silver standard plan. The \texttt{SG} module is trained on \(([x; b],s)\) tuples where \([x;b]\) represents the input document enriched with plan questions (we concatenate input document~$x$ with plan~$b$).  At test time, the trained \texttt{PG} module is used to predict plan~$\hat{b}$ based on input document ${x}$. Generated questions~$\hat{b}$ are then combined with ${x}$ to form the input $[{x}$; $\hat{b}]$ for the \texttt{SG} module, which produces the final summary on the test set.

\texttt{Plan-Output} trains a single end-to-end model that learns to \emph{jointly} generate  the plan and the summary. This design reduces inference overhead, making it well-suited for real-time applications or resource-constrained environments (since only one inference call is made during testing). In contrast, \texttt{Plan-Input} adopts a modular approach, training the  \texttt{PG} and \texttt{SG} components \emph{separately}. This makes \texttt{Plan-Input} better suited to settings where task requirements evolve over time, allowing for the refinement or replacement of individual module components as needed. 

Note that the original Blueprint approach of \citet{narayan-etal-2023-conditional}, relies on \emph{phrase-level} control (the answers to the questions are mostly named entities) and, as such, cannot be readily used to control the generation of explanations that are typically longer and semantically richer. Moreover, the frequent repetition of entities (i.e.,~the same entity is often mentioned multiple times) adds unnecessary complexity to the planning process, necessitating the use of filtering algorithms to streamline it. In contrast, planning at the \emph{EDU-level} provides a more cohesive structure and reduces redundancy. 

%% file: sections/experiments.tex
\section{Experimental Setup}
\label{Experimental_Settings}

\paragraph{Datasets}
We report results on three lay summarization datasets, namely SciNews \cite{pu2024scinews}, eLife \cite{goldsack-etal-2022-making}, and PLOS \cite{goldsack-etal-2022-making}. Table \ref{tab:datasets_statistics} presents different statistics for these datasets, which vary in terms of size, domain, and summary length. In the table, \textsc{Coverage} measures the extent to which the summary directly uses tokens from the source material, with higher coverage indicating that a greater proportion of the summary's tokens come from the source. \textsc{Density} calculates the average length of the source text segments associated with each token in the summary, where higher density suggests the inclusion of longer continuous text segments in the summary. \textsc{Compression Ratio} represents the ratio between the length of the source document and the summary, with a higher compression ratio signifying a more concise summary.

\begin{table*}[t]
\centering
\scalebox{0.67}{
\tabcolsep=5pt
\begin{tabular}{@{}c c c c c c c c c@{}}
\toprule
\textsc{Dataset} & \textsc{\# Training} & \textsc{\# Validation} & \textsc{\# Test} & \textsc{Avg. Doc Tokens} & \textsc{Avg. Summ Tokens} & \textsc{Coverage} & \textsc{Density} &\textsc{Compression Ratio}\\
\midrule
SciNews & 33,497	& 4,187 & 4,188 & 7,760.90 & 694.80 & 0.74 & 0.94 & 12.71\\
eLife & 4,346 & 241 & 241 & 7,833.14 & 383.02 & 0.82 & 1.77 & 20.52 \\
PLOS & 24,773 & 1,376 & 1,376 & 5,340.58 & 178.66 & 0.07 & 0.90 & 36.06 \\
\bottomrule
\end{tabular}
}
\caption{Descriptive statistics for SciNews, eLife, and PLOS datasets.}
\label{tab:datasets_statistics}
\end{table*}

\paragraph{Automatic Evaluation Metrics} 
We report several complementary metrics aimed at assessing different facets of summary quality. We use \textsc{Rouge} \cite{lin-2004-rouge} and \textsc{BERTScore} \cite{zhang2019bertscore} to evaluate \emph{informativeness} against the human references; we report F1 of Rouge-2 (\textsc{R2}) and Rouge-Lsum (\textsc{RLsum}). We use \textsc{D-SARI} \cite{sun-etal-2021-document} and Flesch Reading Ease Formula \citep[\textsc{FRE},][]{kincaid1975derivation} to measure summary \emph{readability}. \textsc{D-SARI} and \textsc{FRE} scores range from~0 to~100, with higher values indicating easier-to-read material and lower scores reflecting more complex passages. Additionally, we compute \textsc{Exp\-Ratio}, which represents the \emph{proportion of explanatory} EDUs to the total number of EDUs in the generated summary (explanatory and target EDUs always appear in pairs due to the RST structure). A higher \textsc{ExpRatio} suggests the summary contains more explanations. We also report the generated summary \emph{length}, which we calculate as the average number of summary tokens (\textsc{AST}) using \href{https://spacy.io/}{spaCy}.

We assess \emph{factual consistency} between the source document and the generated summary using \textsc{SummaC$_{Conv}$} \cite{laban-etal-2022-summac}, which determines whether summary sentences are entailed by the input. However, human-written lay summaries often include additional external information, which could be mistakenly classified as \textit{hallucinations} when assessed automatically with original \textsc{SummaC$_{Conv}$}. This highlights the limitations of traditional consistency detection metrics for our task, where only extrinsic hallucinations, rather than all additional explanations, should be penalized.

To address this issue, we propose a new \textsc{SummaC} variant. Specifically, for summary sentences with entailment scores below~0.5 (i.e., not supported by the source text), we use the \href{https://wikipedia-api.readthedocs.io/en/latest/}{Wikipedia-API} to retrieve relevant articles and re-evaluate these sentences using \textsc{SummaC$_{Conv}$}. This approach allows us to determine whether sentences unsupported by the source text can be validated using external knowledge bases. If the highest entailment score from the retrieved articles is higher than the original score, it replaces the original score for that sentence. We report both the original \textsc{SummaC$_{Conv}$} and the proposed variant, which we call \textsc{SummaC}$^\ast$. We also leverage \textsc{VeriScore} \cite{song-etal-2024-veriscore} to verify whether the claims in the generated summary align with objective facts by consulting external knowledge bases. For claim extraction and verification, we use \texttt{GPT-4o}.

\paragraph{Model Comparisons} 
We build \mbox{\texttt{Plan-Output}} and \mbox{\texttt{Plan-Input} (\texttt{PG} and \texttt{SG})} on top of the \mbox{\texttt{Mistral-7B-Instruct-v0.3}}, which we fully fine-tune on the above datasets (enriched with plan annotations). We compare our models against the following \texttt{Mistral} configurations: a) zero-shot setting (\texttt{Mistral}$_{ZS}$); b)~in-context learning with one randomly selected demonstration from the training split (\texttt{Mistral}$_{ICL}$); and c)~full parameter fine-tuning without planning (\texttt{Mistral}$_{FT}$). 

We also re-implement the best-performing multi-task blueprint model (\texttt{Blueprint}$_{MT}$) from \citet{narayan-etal-2023-conditional}, which is optimized for two tasks: 1)~question generation given source document and said answer, or 2)~summary generation given the same document and its question-answer plan. We use \texttt{GPT-4o} to identify answers (while the original \texttt{Blueprint} model extracts answers using \href{https://spacy.io/}{spaCy}, we employ \texttt{GPT-4o} which leads to better performance -- on average 2.21\% higher \textsc{RLsum} across datasets) and generate corresponding questions and fine-tune with the same backbone \texttt{Mistral} model on datasets enriched with question-answer plans.\footnote{Prompts are offered in Appendix Figures \ref{prompt_zero_shot_learning}, \ref{prompt_in_context_learning}, \ref{prompt_named_entities} and \ref{prompt_fine_plan_tuning_learning}.}

All comparison models follow identical hyperparameter settings detailed in Appendix~\ref{hyperparameter_settings}. We also include results with \texttt{GPT-4o} \cite{achiam2023gpt} in zero-shot (\texttt{GPT-4o}$_{ZS}$) and in-context learning (\texttt{GPT-4o}$_{ICL}$) settings; we compare against the state-of-the-art (SOTA) methods on each dataset (results are directly taken from respective publications). For a fair comparison, we use the same prompt and/or selected sample for both \texttt{Mistral}$_{ZS/ICL}$ and \texttt{GPT-4o}$_{ZS/ICL}$. After inference, we discard all generated plans, retaining only the summaries for automatic evaluation.

\begin{table*}[t]
\centering
\scalebox{0.6}{
\begin{threeparttable}
\begin{tabular}{c l c c c c c c c c c c c c}
\toprule
\textsc{Data} & \textsc{Model} & \textsc{R2} & \textsc{RLsum} & \textsc{BERTscore} & \textsc{D-SARI} & \textsc{FRE} & \textsc{ExpRatio} & \textsc{AST} & \textsc{SummaC/SummaC$^\ast$} & \textsc{VeriScore} \\
\hline
\multirow{10}*{\rotatebox[origin=c]{90}{SciNews}} 
& \texttt{Mistral}$_{ZS}$ & \hspace*{1.3ex}7.07 & 37.02 & 57.21 & 14.12 & 27.41 & 10.45 & 554.29 & 46.11/52.65 & 0.43 \\
~ & \texttt{Mistral}$_{ICL}$ & \hspace*{1.3ex}7.11 & 37.31 & 57.57 & 14.65 & 31.93 & 9.87 & 602.33 & 50.12/59.27 & 0.48\\
~ & \texttt{GPT-4o}$_{ZS}$ & 12.79 & 40.51 & 58.15 & 20.15 & 33.05 & 11.73 & 611.09 & 58.21/64.96 & 0.56 \\
~ & \texttt{GPT-4o}$_{ICL}$ & 13.22 & 40.73 & 58.17 & 20.33 & 38.91 & 11.90 & 634.76 & 60.04/66.74 & 0.52\\
\cdashline{2-11}
~ & \texttt{Mistral}$_{FT}$ & 15.56 & 46.12 & 64.38 & 30.11 & 38.25 & 13.61 & 669.27 & 64.10/69.73 & 0.56\\
~ & \texttt{Blueprint}$_{MT}$ & 15.62 & 46.10 & 64.33 & 32.35 & 40.01 & 15.03 & 688.93 & 72.35/75.18 & 0.62\\
~ & \texttt{Plan-Output} & 15.73 & 46.30 & \cellcolor{Coral}\textbf{65.34} & 36.23 & 42.25 & 17.51  & 685.33 & 72.29/75.03 & 0.67\\
~ & \texttt{Plan-Input} & \cellcolor{Coral}\hspace{2ex}\textbf{15.88}$^\dag$$^\ddag$ & \cellcolor{Coral}\hspace{2ex}\textbf{46.41}$^\dag$$^\ddag$ & \hspace{2ex}65.32$^\dag$$^\ddag$ & \cellcolor{Coral}\hspace{2ex}\textbf{37.18}$^\dag$$^\ddag$ & \cellcolor{Coral}\hspace{2ex}\textbf{43.11}$^\dag$$^\ddag$ & \cellcolor{Coral}\hspace{2ex}\textbf{17.68}$^\dag$$^\ddag$ & \cellcolor{Coral}\hspace{2ex}\textbf{692.14}$^\dag$$^\ddag$ & \cellcolor{Coral}\hspace{2ex}\textbf{72.40}$^\dag$/\cellcolor{Coral}\textbf{75.38}$^\dag$$^\ddag$ & \cellcolor{Coral}\hspace{2ex}\textbf{0.71}$^\dag$$^\ddag$ \\
~ & \citet{pu2024scinews} & 14.92 & 45.19 & 62.80 & --- & ---&  --- & --- & --- & --- \\
\cdashline{2-11}
~ & Reference Summary & --- & --- & --- & --- & 42.23 & 18.91 & 694.80 & 45.08/81.94 & 0.81\\
\midrule
\midrule
\multirow{12}*{\rotatebox[origin=c]{90}{eLife}}
& \texttt{Mistral}$_{ZS}$ & \hspace*{1.3ex}8.90 & 36.19 & 61.38 & 18.28 & 28.43 & 10.29 & 301.16 & 45.45/56.68 & 0.39\\
~ & \texttt{Mistral}$_{ICL}$ & \hspace*{1.3ex}8.97 & 36.26 & 61.40 & 20.07 & 33.60 & 10.30 & 333.27 & 47.42/58.98 & 0.45\\
~ & \texttt{GPT-4o}$_{ZS}$ & 11.15 & 44.29 & 64.34 & 24.57 & 38.17 & 11.10 & 329.39 & 47.30/58.46 & 0.56\\
~ & \texttt{GPT-4o}$_{ICL}$ & 11.31 & 44.38 & 63.77 & 25.52 & 44.19 & 11.22 & 368.42 & 50.92/60.07 & 0.60\\
\cdashline{2-11}
~ & \texttt{Mistral}$_{FT}$ & 14.40 & 47.68 & 86.87 & 32.60 & 48.42 & 13.56 & 377.59 & 58.31/62.33 & 0.58\\
~ & \texttt{Blueprint}$_{MT}$ & 14.85 & 48.09 & 87.22 & 35.31 & 50.33 & 14.51 & 375.47 & 60.31/65.44 & 0.62\\
~ & \texttt{Plan-Output} & 14.91 & 48.27 & 87.57 & 38.77 & \cellcolor{Coral}\textbf{56.76} & 17.71 & 377.61 & 61.33/68.21 & 0.70\\
~ & \texttt{Plan-Input}& \cellcolor{Coral}\hspace{2ex}\textbf{15.11}$^\dag$$^\ddag$ & \cellcolor{Coral}\hspace{1.8ex}\textbf{48.64}$^\dag$$^\ddag$ & \cellcolor{Coral}\hspace{1.8ex}\textbf{87.92}$^\dag$$^\ddag$ & \cellcolor{Coral}\hspace{2ex}\textbf{39.22}$^\dag$$^\ddag$ & \hspace{2ex}55.48$^\dag$$^\ddag$ & \cellcolor{Coral}\hspace{2ex}\textbf{17.73}$^\dag$$^\ddag$ & \cellcolor{Coral}\hspace{2ex}\textbf{380.39}$^\dag$$^\ddag$ & \cellcolor{Coral}\hspace{2ex}\textbf{61.35}$^\dag$$^\ddag$/\cellcolor{Coral}\textbf{68.28}$^\dag$$^\ddag$ & \cellcolor{Coral}\hspace{2ex}\textbf{0.75}$^\dag$$^\ddag$\\
~ & \citet{goldsack-etal-2023-enhancing} & 14.24 & 45.71 & 85.40 & --- & --- & --- & --- & --- & ---\\
~ & \citet{pu-demberg-2024-rstlora} & 14.92 & 48.21 & 87.81 & --- & --- & --- & ---  & --- & ---\\
~ & \citet{Zhang2024ATLASIL} & 12.57 & 44.14 & 85.20 & --- & --- & --- & --- & --- & ---\\
\cdashline{2-11}
~ & Reference Summary & --- & --- & --- & --- & 51.83 & 19.02 & 383.02 & 49.38/82.28 & 0.83\\
\midrule
\midrule
\multirow{11}*{\rotatebox[origin=c]{90}{PLOS}}
& \texttt{Mistral}$_{ZS}$ & \hspace*{1.3ex}6.43 & 31.42 & 63.24 & 19.09 & 29.70 & 11.03 & 130.14 & 45.83/67.47 & 0.41\\
~ & \texttt{Mistral}$_{ICL}$ & \hspace*{1.3ex}6.50 & 31.66 & 63.31 & 21.27 & 35.93 & 11.19 & 149.30 & 49.02/68.87 & 0.44\\
~ & \texttt{GPT-4o}$_{ZS}$ & 11.57 & 35.39 & 64.22 & 21.06 & 32.69 & 12.31 & 155.25 & 55.23/70.79 & 0.52\\
~ & \texttt{GPT-4o}$_{ICL}$ & 11.72 & 35.40 & 64.50 & 23.74 & 39.44 & 12.31 & 152.38 & 58.02/73.03 & 0.50\\
\cdashline{2-11}
~ & \texttt{Mistral}$_{FT}$ & 14.31 & 40.22 & 87.69 & 32.39 & 36.45 & 13.70 & 158.93 & 60.02/75.25 & 0.57\\
~ & \texttt{Blueprint}$_{MT}$  & 14.20 & 41.35 & 88.04 & 33.48 & 36.04 & 14.67 & 166.54 & 62.22/77.67 & 0.63\\
~ & \texttt{Plan-Output} & 15.28 & 41.33 & 88.17 & 39.22 & 40.01 & \cellcolor{Coral}\textbf{18.12} & 167.77 & 62.25/77.79 & \cellcolor{Coral}\textbf{0.74}\\
~ & \texttt{Plan-Input} & \cellcolor{Coral}\hspace{2ex}\textbf{15.72}$^\dag$$^\ddag$ & \cellcolor{Coral}\hspace{2ex}\textbf{41.64}$^\dag$$^\ddag$ & \cellcolor{Coral}\hspace{1.6ex}\textbf{88.22}$^\dag$$^\ddag$ & \cellcolor{Coral}\hspace{2ex}\textbf{40.07}$^\dag$$^\ddag$ & \cellcolor{Coral}\hspace{2ex}\textbf{40.27}$^\dag$$^\ddag$
& \hspace{2ex}17.65$^\dag$$^\ddag$ & \cellcolor{Coral}\hspace{2ex}\textbf{171.12}$^\dag$$^\ddag$ & \cellcolor{Coral}\hspace{2ex}\textbf{62.27}$^\dag$/\cellcolor{Coral}\textbf{77.95}$^\dag$$^\ddag$ & \hspace{2ex}0.72$^\dag$$^\ddag$\\
~ & \citet{goldsack-etal-2022-making} & 13.52 & 38.63 & --- & --- & --- & --- & --- & --- & ---\\
~ & \citet{Zhang2024ATLASIL} & 12.33 & 40.60 & 85.70 & --- & ---  & --- & --- & --- & ---\\
\cdashline{2-11}
~ & Reference Summary & --- & --- & --- & --- & 30.08 & 18.19 & 178.66 & 53.43/84.71 & 0.87\\
\bottomrule
\end{tabular}
\end{threeparttable}
}
\caption{Model performance on three lay summarization datasets. Bold numbers represent the best results achieved by the models in each test set, excluding human results from the comparison. \textsc{ExpRatio} is the number of explanatory EDUs over all EDUs in summaries. Symbols~$^\dag$ and~$^\ddag$  denote that \texttt{Plan-Input} is statistically significant 
(\mbox{$p<0.05$}) against  \texttt{Mistral}$_{FT}$ and \texttt{Blueprint}$_{MT}$ using paired t-test, respectively.}
\label{tab:model_performance}
\end{table*}

\section{Results}
\label{general_results}
Our results are summarized in Table~\ref{tab:model_performance}, which consists of three main blocks corresponding to different datasets. Within each block, we compare models in zero-shot and in-context learning settings against fine-tuned systems and, where applicable, against the human reference summaries. 

We find the performance of fine-tuned models to be superior. Across all metrics and datasets, \texttt{Plan-Input} emerges as the top-performing model, followed closely by \texttt{Plan-Output} (see \textsc{R2} and \textsc{RLsum}). Recall that \texttt{Plan-Input} is a pipeline system where the plan and summary generation stages are trained separately, which we hypothesize makes the summarization task easier. 

In comparison to the related \texttt{Blueprint}$_{MT}$ and other comparison models, our two discourse-driven systems produce summaries that are easier to read (see \textsc{FRE} and \textsc{D-SARI}), contain more explanations (see \textsc{Exp\-Ratio}), are closer to human-level performance (see Reference Summary), and align more closely with human summaries in terms of length (\textsc{AST}). Since \texttt{Blueprint}$_{MT}$ relies on entities and phrases to generate plans, it lacks diversity in the types of questions it predicts; in fact, \textit{what} and \textit{who} questions are the greatest majority with an average of~82.2\% across datasets. In contrast, our models produce a more balanced distribution, with \textit{what} and \textit{who} questions making up 23.2\% of the plan, while \textit{how} and \textit{why} questions account for an average of~66.4\%.

As human-authored summaries often include content that is not directly mentioned in the source document \cite{cao-etal-2022-hallucinated}, we observe relatively low (original) \textsc{Summac} scores. However, when external knowledge bases are considered, we find that human summaries exhibit the lowest level of extrinsic hallucinations. \textsc{VeriScore} results further corroborate the observation that model-generated summaries lag behind human-authored ones regarding factual consistency. Nevertheless, fine-tuned models are overall more consistent, with plan-based models having an advantage against \texttt{Mistral}$_{FT}$ and state-of-the-art systems. Both \texttt{Plan-Input} and \texttt{Plan-Output} achieve the highest consistency scores across datasets and metrics, indicating that planning at the EDU-level not only enhances consistency compared to phrase-level planning but also enables the generation of summaries with more elaborations while maintaining lower levels of hallucinations.

\paragraph{Impact of Plan Generation Ablation}
\label{ablation_results}

Next, we ablate our plan generation strategy more closely, comparing it to naive alternatives. We apply \texttt{Plan-Input} as our main model, except when indicated otherwise. 

\begin{enumerate}[itemsep=1pt,topsep=1pt,parsep=1pt]
\item Lead-3$_Q$ and Lead-K$_Q$: Generate plans based on the first 3 or first $K$~EDUs in the reference summary.
\item Tail-3$_Q$ and Tail-K$_Q$: Generate plans based on the last 3 or last $K$~EDUs in  the reference summary.
\item Random-3$_Q$ and Random-K$_Q$: Generate plans based on 3 random or $K$~random EDUs in the reference summary. 
\item All-EDUs$_Q$: Generate plans based on each of the EDUs in the reference summary.
\item NonExp-EDUs$_Q$: Generate plans based on each of the non-explanatory EDUs in the reference summary.
\item \texttt{PG} for zero-shot learning: Train a \texttt{PG} model and use the plan questions it generates as instructions to \texttt{Mistral} in a zero-shot setting.\footnote{The prompt can be found in Appendix Figure \ref{prompt_fine_plan_tuning_learning}.}
\end{enumerate}

Here, \(K\) denotes the average number of explanatory EDUs parsed in training set summaries, rounded up to the nearest whole number (\(K=8\) for SciNews, \(K=4\) for eLife, and \(K=2\) for PLOS). For the first four experiments, we do not differentiate between explanatory and non-explanatory EDUs. Each selected EDU serves as the target sentence; all preceding EDUs are considered context and are used to generate the corresponding plan questions. For instance, in the All-EDUs$_Q$ experiment (all sentences from the reference summary are involved in the plan generation, rather than being limited to explanatory EDUs), the first sentence is treated as a target EDU (since there are no preceding EDUs, there is no context in this case) and the first plan question is generated. Next, the second EDU is treated as a target EDU, and the first EDU is used as context, and so on.  

As shown in Table~\ref{tab:ablation_study}, Lead-3$_Q$, Lead-K$_Q$, and All-EDUs$_Q$ strategies demonstrate marginal improvements over \texttt{Mistral}$_{FT}$ across all datasets, while all other strategies result in decreased performance. All strategies in Table~\ref{tab:ablation_study} are worse than \texttt{Blueprint}$_{MT}$, \texttt{Plan-Input}, and \texttt{Plan-Output}, in terms of summary quality and factual consistency. This implies that heuristic ablations are less effective for the generation of plan questions. Results for NonExp-EDUs$_Q$ and All-EDUs$_Q$ also verify that modeling and generating plan questions for explanatory sentences is more valuable for lay summarization than modeling and generating plans for other sentences. Results for \texttt{Mistral}$_{ZS+PG}$ further show that a robust plan generator improves performance, even in zero-shot settings.

\begin{table}[t]
\centering
\scalebox{0.9}{
\tabcolsep=3pt
\begin{threeparttable}
\begin{tabular}{c l c c c}
\toprule
\textsc{Data} & \textsc{Model} & \textsc{R2} & \textsc{RLsum} & \textsc{VeriScore} \\
\hline
\multirow{9}*{\rotatebox[origin=c]{90}{SciNews}} 
~ & \texttt{Mistral}$_{FT}$ & 15.56 & 46.12 & 0.56 \\
\cdashline{2-5}
~ & Lead-3$_Q$ & 15.47 & 46.12 & 0.59 \\
~ & Lead-K$_Q$ & 15.50 & 46.15 & 0.52 \\
~ & Tail-3$_Q$ & 15.41 & 45.80 & 0.50 \\
~ & Tail-K$_Q$ & 15.40 & 45.85 & 0.50 \\
~ & Random-3$_Q$ & 15.35 & 45.88 & 0.48 \\
~ & Random-K$_Q$ & 15.33 & 45.94 & 0.47 \\
~ & All-EDUs$_Q$ & 15.48 & 46.11 & 0.52 \\
~ & NonExp-EDUs$_Q$ & 15.42 & 46.03 & 0.49 \\
\cdashline{2-5}
~ & \texttt{Mistral}$_{ZS+PG}$ & 7.15 & 37.11 & 0.46 \\
\midrule
\midrule
\multirow{9}*{\rotatebox[origin=c]{90}{eLife}}
~ & \texttt{Mistral}$_{FT}$ & 14.40 & 47.68 & 0.58 \\
\cdashline{2-5}
~ & Lead-3$_Q$ & 14.68 & 48.14 & 0.59 \\
~ & Lead-K$_Q$ & 14.71 & 48.11 & 0.60 \\
~ & Tail-3$_Q$ & 14.25 & 47.85 & 0.54 \\
~ & Tail-K$_Q$ & 14.30 & 47.74 & 0.54 \\
~ & Random-3$_Q$ & 14.65 & 48.05 & 0.56 \\
~ & Random-K$_Q$ & 14.62 & 48.00 & 0.55 \\
~ & All-EDUs$_Q$ & 14.62 & 48.10 & 0.56 \\
~ & NonExp-EDUs$_Q$ & 14.56 & 48.02 & 0.52 \\
\cdashline{2-5}
~ & \texttt{Mistral}$_{ZS+PG}$ & 10.11 & 38.69 & 0.42 \\
\midrule
\midrule
\multirow{9}*{\rotatebox[origin=c]{90}{PLOS}}
~ & \texttt{Mistral}$_{FT}$ & 14.31 & 40.22 & 0.57 \\
\cdashline{2-5}
~ & Lead-3$_Q$ & 14.69 & 40.66 & 0.59 \\
~ & Lead-K$_Q$ & 14.57 & 40.59 & 0.56 \\
~ & Tail-3$_Q$ & 13.78 & 40.12 & 0.52 \\
~ & Tail-K$_Q$ & 13.72 & 40.18 & 0.51 \\
~ & Random-3$_Q$ & 14.21 & 40.26 & 0.54 \\
~ & Random-K$_Q$ & 14.14 & 40.20 & 0.53 \\
~ & All-EDUs$_Q$ & 14.41 & 40.33 & 0.55 \\
~ & NonExp-EDUs$_Q$ & 14.37 & 40.33 & 0.54 \\
\cdashline{2-5}
~ & \texttt{Mistral}$_{ZS+PG}$ & 6.74 & 32.44 & 0.48 \\
\bottomrule
\end{tabular}
\end{threeparttable}
}
\caption{Comparison of different plan generation strategies using \texttt{Plan-Input} and \texttt{Mistral} in fine-tuned and zero-shot settings.}
\label{tab:ablation_study}
\end{table}

%% file: sections/analysis.tex
\paragraph{Impact of Discourse Parser}
\label{parser}

We next investigate the extent to which the choice of RST parser influences summary output quality. Specifically, we compare the \texttt{DMRST} parser, which has been trained on the RST treebank, against results derived from \texttt{GPT-4o} and \texttt{Mistral} in zero-shot settings (the instruction is detailed in Appendix Figure~\ref{GPT-4o/mistral_as_RST_Parser}). For \texttt{GPT-4o} and \texttt{Mistral}, we use these models to directly extract explanatory sentences and their target sentences, rather than performing RST parsing, as RST parsing is a more complex task.\footnote{The performance is better in the former way—on average, the \textsc{RLsum} score is 7.82\% higher than the latter across datasets when applied \texttt{Mistral} under zero-shot setting.} Following \citet{stede2017annotation}, we implement a rule-based method (RB) for extracting explanations (detailed in Appendix~\ref{rule_based_method}). We also use the RST-Coref parser \cite{guz-carenini-2020-coreference} and LLaMA-based RST parser \cite{maekawa-etal-2024-obtain} for comparison. To simulate parser instability, we design a random replacement (RR) method, where explanatory EDUs identified by \texttt{DMRST} are randomly replaced with non-explanatory or non-target EDUs. The number of replacements varies from one to the total number of explanatory EDUs per article, mimicking parser inaccuracy in identifying explanatory content. Finally, we introduce full random replacement (FRR), a variant representing the worst-case scenario of parser inaccuracy, where all explanatory EDUs are replaced with randomly selected non-explanatory or non-target EDUs. All these models are trained using \texttt{Plan-Input} as the default model.

\begin{figure}[t]
  \centering
  \includegraphics[width=0.5\textwidth]{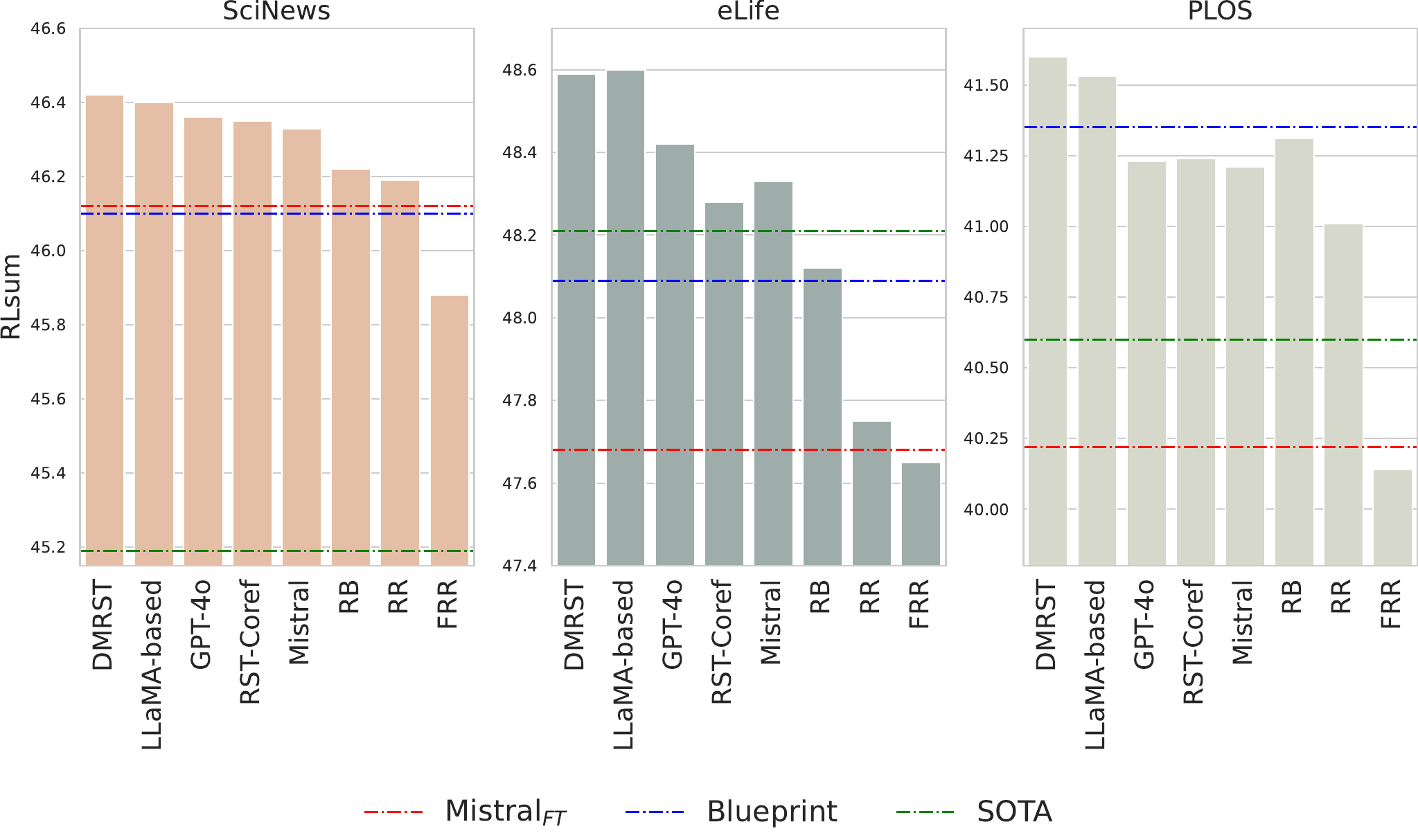}
  \caption{Summary quality as a function of different RST parsers.}
  \label{fig:parser_capability_impact}
\end{figure}

Figure~\ref{fig:parser_capability_impact} shows that summarization performance improves with parsing accuracy (see \texttt{DMRST} vs.~RR/FRR). \texttt{DMRST} consistently outperforms more naive methods based on \mbox{\texttt{GPT-4o}} and \texttt{Mistral}, as well as the RST-Coref parser and the rule-based approach. Its performance is comparable to the LLaMA-based RST parser, but it has a smaller parameter size and incurs lower computational costs in parsing. We also find that RST-based models (irrespectively of the parser at hand) outperform the baseline fine-tuned (\texttt{Mistral}$_{FT}$) and SOTA models (the performance of the previous best-published model on \textsc{RLSum} for each dataset in Table \ref{tab:model_performance}), and frequently surpass the \texttt{Blueprint}$_{MT}$ model, as illustrated by the red, green, and blue dashed lines, respectively.

\paragraph{Impact of Plan Quality}
To evaluate how the quality of plan questions affects model performance, we employ \texttt{GPT-4o} in a zero-shot setting as a question generator (see Section~\ref{sec:plan_generation}). In addition, we apply the \texttt{Mistral} model and the recent question generation model (\texttt{RAST}), developed by \citet{gou-etal-2023-diversify}. Moreover, we implement another random replacement (RR) method, where questions generated by \texttt{GPT-4o} are randomly substituted with irrelevant ones.\footnote{The prompt can be found in Appendix Figure~\ref{Irrelevant_Question_Generation}.} The number of questions replaced varies from one to all for each article. Finally, we introduce another full random replacement (FRR), where questions generated by \texttt{GPT-4o} are all replaced with random irrelevant questions. Throughout these experiments, \texttt{Plan-Input} remains the default model.

\begin{figure}[t]
  \centering
  \includegraphics[width=0.5\textwidth]{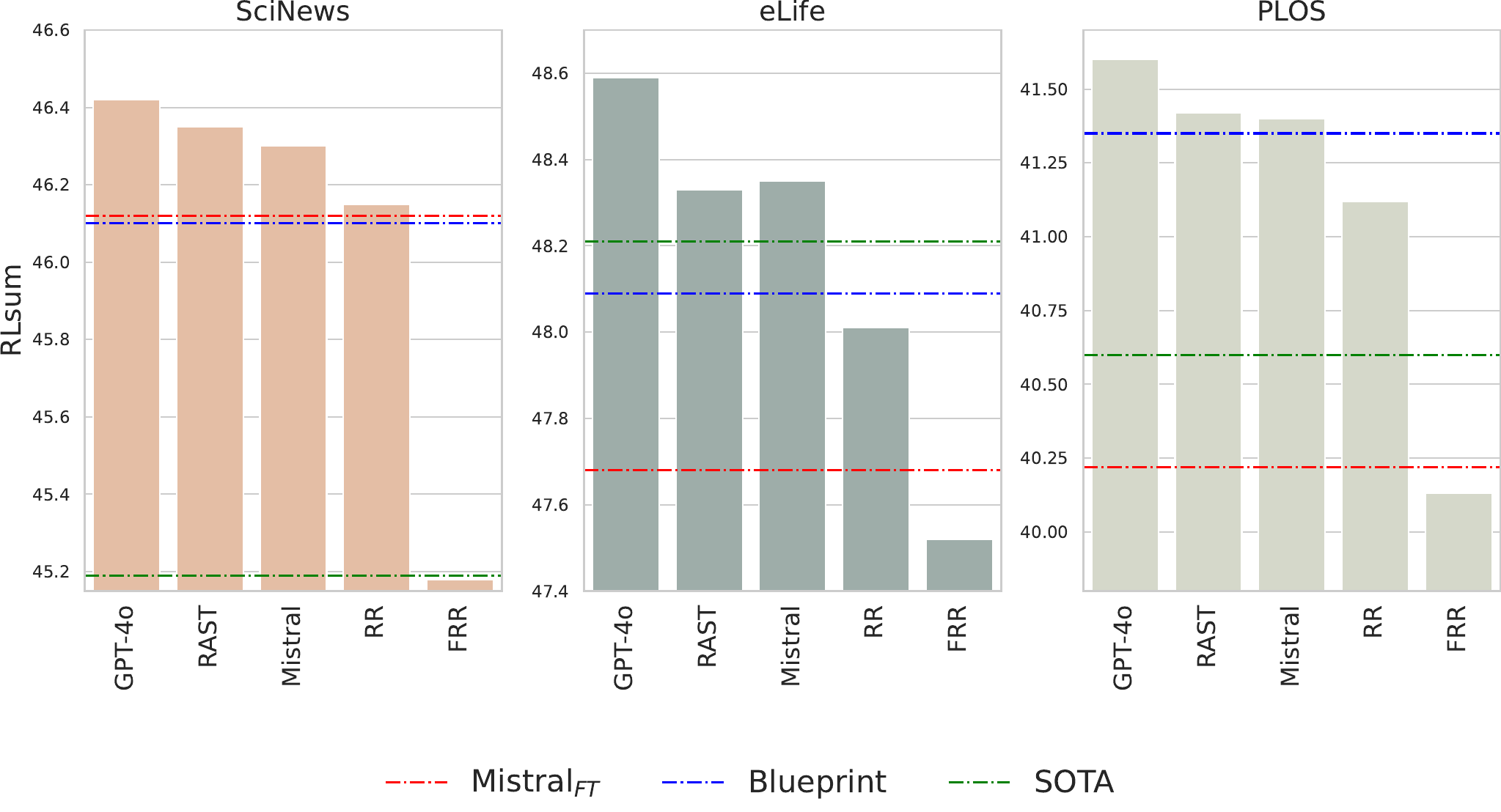}
  \caption{Summary quality as a function of different question generation methods.}
  \label{fig:question_quality_impact}
\end{figure}

Figure~\ref{fig:question_quality_impact} reveals that plan questions have a direct impact on summary quality (compare \texttt{GPT-4o} vs. RR/FRR). Our method exhibits a certain degree of robustness, as it performs reasonably well even when the questions contain some degree of noise (RR vs. FRR). In the FRR condition,  model performance degrades substantially, which suggests that relevant plan questions are instrumental in improving summary quality. Although questions generated by \texttt{Mistral} and \texttt{RAST} lead to lower performance gains compared to \texttt{GPT-4o}, these open-source models offer a cost-effective alternative. Despite not reaching the highest performance, they still surpass the previous state-of-the-art results.

\paragraph{Impact of Plan Composition}
An advantage of plan-based models is their capacity to directly control the content of the output summary by simply altering the plan that gives rise to it. In this section, we investigate the extent to which changes in the formation of the plan questions can be observed in the generated summaries.\footnote{We modify the plan generated by the model and evaluate whether our modifications have any bearing on the output.} To this end, we conduct a controlled generation experiment. We investigate whether our models can be guided to avoid generating these particular types of explanations by eliminating questions from the plan that correspond to specific types of explanatory EDUs.

Specifically, we remove from the plan questions corresponding to \emph{Background}, \emph{Comparison}, \emph{Elaboration}, and \emph{Explanation} (EDUs are identified by the \texttt{DMRST} parser). Table~\ref{tab:type_control} reports the proportion of explanatory EDUs in the output summaries for our two models when no manipulation takes place. These results are compared with summaries generated from plans in which specific types of explanations have been deliberately removed. By deleting plan questions related to certain types of explanations, we evaluate the models' capacity to refrain from generating particular types of EDUs. Results presented in Table~\ref{tab:type_control} confirm that our discourse-driven planning approach is sufficiently expressive to control the explanations and their discourse function in the summary. For both \texttt{Plan-Input} and \texttt{Plan-Output}, we observe a close alignment between the questions in the plan and the types of explanations produced in the summary.

\begin{table}[t]
\centering
\scalebox{0.6}{
\tabcolsep=4pt
\begin{tabular}{c c ccc ccc c c}
\toprule
\multirow{2}{*}{\textsc{Data}} & \multirow{2}{*}{\textsc{Relation}}& \multicolumn{3}{c}{\texttt{Plan-Output}} & & \multicolumn{3}{c}{\texttt{Plan-Input}} \\ 
\cmidrule(lr){3-5} \cmidrule(lr){7-9}
 & & \textsc{NoDel} & \textsc{Del} & $\Delta$\textsc{RLSum} & & \textsc{NoDel} & \textsc{Del} & $\Delta$\textsc{RLSum} \\ 
\midrule
\multirow{4}*{\rotatebox[origin=c]{90}{SciNews}} 
& {Background} & 4.04 & 0.98 & 0.45 & & 4.18 & 1.07 & 0.42 \\ 
& {Comparison} & 2.21 & 0.21 & 0.38 & & 2.60 & 0.33 & 0.39 \\
& {Elaboration} & 6.47 & 1.25 & 1.56 & & 6.39 & 1.20 & 1.50 \\ 
& {Explanation} & 7.09 & 1.69 & 1.62 & & 7.88 & 2.11 & 1.56 \\ 
\midrule
\multirow{4}*{\rotatebox[origin=c]{90}{eLife}} 
& {Background} & 4.38 & 1.04 & 0.27 & & 4.24 & 1.02 & 0.24 \\ 
& {Comparison} & 2.40 & 0.17 & 0.13 & & 2.59 & 0.19 & 0.15 \\
& {Elaboration} & 5.46 & 1.16 & 0.46 & & 5.42 & 1.15 & 0.40 \\ 
& {Explanation} & 6.07 & 1.72 & 0.58 & & 6.11 & 1.81 & 0.52 \\ 
\midrule
\multirow{4}*{\rotatebox[origin=c]{90}{PLOS}} 
& {Background} & 4.33 & 0.97 & 0.20 & & 4.09 & 0.92 & 0.22 \\ 
& {Comparison} & 2.02 & 0.09 & 0.11 & & 2.12 & 0.13 & 0.13 \\
& {Elaboration} & 5.25 & 1.13 & 0.26 & & 5.32 & 1.15 & 0.25 \\ 
& {Explanation} & 6.42 & 1.52 & 0.48 & & 6.72 & 1.49 & 0.42 \\ 
\bottomrule
\end{tabular}
}
\caption{Proportion of explanatory EDUs in predicted summaries for models with (\textsc{Del}) and without (\textsc{NoDel}) deletions in the plan. In the \textsc{NoDel} setting, the model regulates the proportion of explanatory content in the plan and summary. In the \textsc{Del} setting, the plan is manipulated by removing specific explanations. $\Delta$\textsc{RLsum} = \textsc{RLsum} before plan deletion $-$ \textsc{RLsum} after plan deletion.}
\label{tab:type_control}
\end{table}

Amongst explanatory relations, \textit{Elaboration} and \textit{Explanation} emerge as the most significant contributors to overall summary quality. In the \texttt{Plan-Input} model, the deletion of \textit{Elaboration} explanations consistently results in lower \textsc{RLsum} across datasets (e.g.,~a drop from 46.41 to 44.91 in SciNews and from 48.64 to 48.24 in eLife). In contrast, removing \textit{Comparison} explanations has a minor effect on summary quality, as reflected in the relatively stable \textsc{RLsum} scores across datasets (e.g.,~41.64 vs.~41.51 in PLOS and 48.64 vs.~48.49 in eLife). In the \textsc{NoDel} condition, we observe that \textit{Elaboration} and \textit{Explanation} relations are more prominent in the summary texts, which suggests that removing them will be more detrimental (e.g.,~compared to \emph{Background} or \emph{Comparison}).

\section{Human Evaluation}
\label{human_evaluation}

To alleviate the limitations of automated metrics and more profoundly analyze the quality of model-generated summaries, we randomly select 5 instances from each of the SciNews, eLife, and PLOS datasets (a total of 15) for human evaluation. The evaluators are postgraduate or doctoral students specializing in Computer Science or Computational Linguistics with advanced proficiency in English. They are compensated at our university's standard hourly rate, and remain blind to the origin of each summary during the evaluation process. In this study, we compare \texttt{Mistral}$_{FT}$, \texttt{Blueprint}$_{MT}$, \texttt{Plan-Input} (our best performing model), \texttt{GPT-4o}$_{ZS}$, and human-authored reference summaries. Each summary is independently rated by three different evaluators, resulting in a total of~225 $(15 \times 5 \times 3)$ evaluation samples.

Our judges rate each summary on a 1 to 5 Likert scale across the dimensions of Faithfulness, Relevance, Informativeness, Accessibility, Explanation Accuracy, and Explanation Usefulness, with higher scores indicating better quality. They are also tasked with ranking the summaries from highest to lowest based on their relative quality within the set of candidates. Raters highlight explanations while reading the summary and the source document. To assess the explanations' accuracy and usefulness, judges are allowed to consult external knowledge sources, such as books, academic papers, and Wikipedia, but not any AI tools. When they are not able to judge the accuracy or usefulness of explanations, raters are asked to skip them and flag them for expert review. We provide detailed evaluation guidelines in Appendix~\ref{human_evaluation_guideline} and discuss cases flagged for expert review in Appendix~\ref{Expert_Results}.

\begin{figure}[t]
  \centering
  \includegraphics[width=0.45\textwidth]{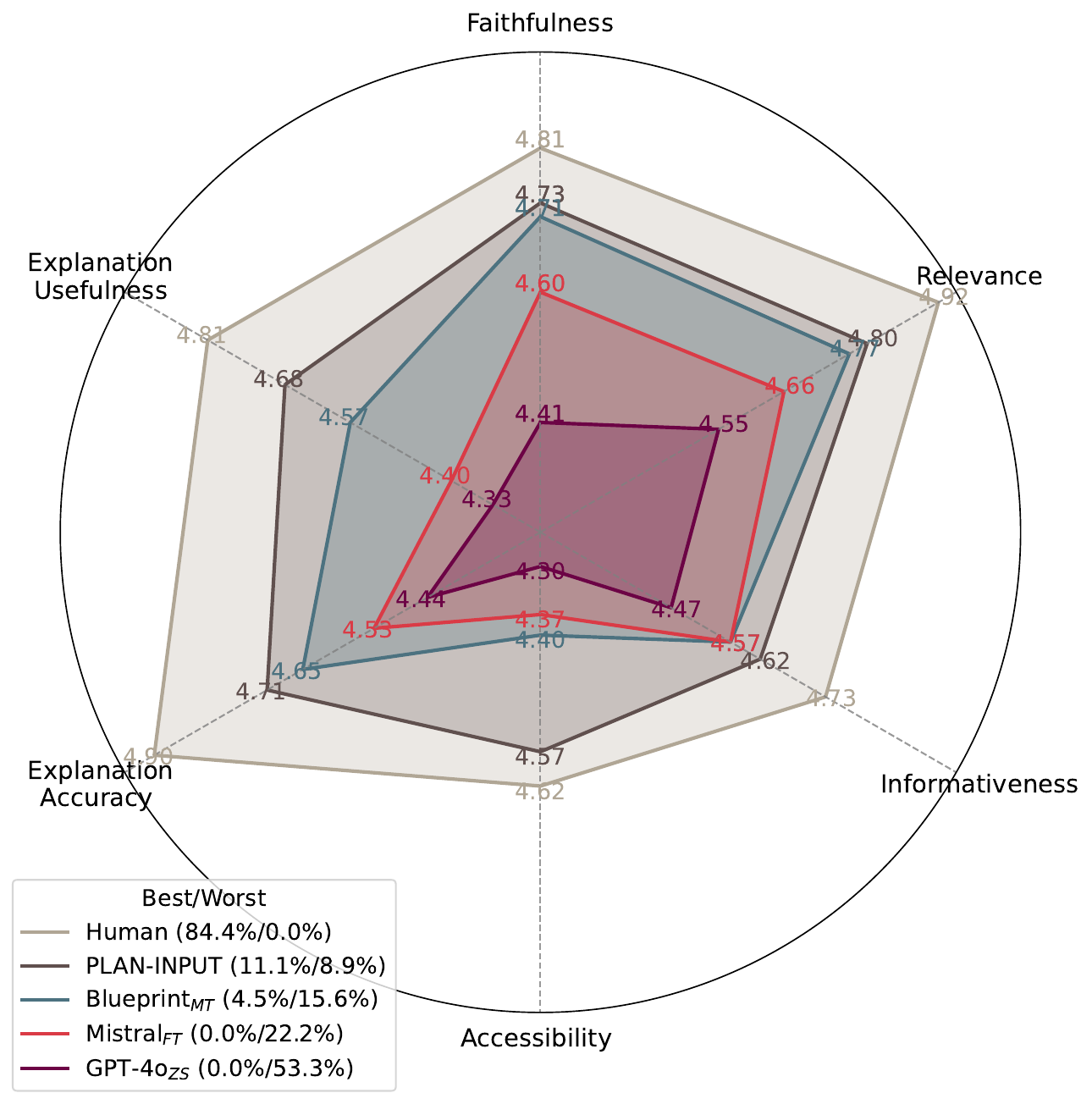}
  \caption{Human evaluation results along different dimensions of summary quality.}
  \label{fig:human_evaluation}
 \end{figure}

Figure~\ref{fig:human_evaluation} reports the performance of each model across all summary samples, as well as the proportion of times a model is rated best (or worst). Fleiss' Kappa scores for Faithfulness (\mbox{$\kappa$= 0.712}), Relevance (\mbox{$\kappa$ = 0.835}), Informativeness (\mbox{$\kappa$ = 0.673}), Accessibility (\mbox{$\kappa$ = 0.707}), Explanation Accuracy (\mbox{$\kappa$ = 0.604}), and Explanation Usefulness (\mbox{$\kappa$ = 0.633}) indicate substantial inter-rater agreement, with an average of \mbox{$\kappa$ = 0.694}. Overall, we observe that human-written summaries outperform all neural summarization models in terms of quality, with particularly pronounced differences in the usefulness and accuracy of explanations. Notably, human-written summaries also perform best in terms of faithfulness.

Among the four neural models under consideration, \texttt{GPT-4o}$_{ZS}$ performs worst, while \texttt{Mistral}$_{FT}$ also obtains low results, with a 22.2\% likelihood of being rated as worst. Models based on planning (\texttt{Blueprint}$_{MT}$ and \texttt{Plan-Input}) outperform \texttt{Mistral}$_{FT}$, with \texttt{Plan-Input} being superior across all metrics. On certain criteria, such as accessibility, \texttt{Plan-Input} is on par with human summaries. Compared to other neural summarization systems, it is also more likely to generate high-quality summaries. 

\section{LLM-as-Judge Evaluation}

Due to the considerable length of the source documents and their summaries, it is not feasible to conduct extensive manual evaluations. Therefore, we also use an LLM-based evaluator \cite{liusie-etal-2024-llm, zheng2023judging, liu2025talkaboutvideototextsummarization} to perform large-scale comparisons of system outputs. We use the prompt targeting the same dimensions of summary quality adopted in our human evaluation and use \texttt{GPT-4o} as our evaluator (following the hyperparameter settings outlined in Appendix~\ref{hyperparameter_settings}). To avoid potential bias from previous interactions, we reset the conversation history before each query, making no further adjustments.

Firstly, we validate the agreement between \texttt{GPT-4o} and human ratings using the same set of 15 samples originally employed for human evaluation.  We obtain a single rating per sample by averaging the scores of individual participants. We compute Fleiss' Kappa to measure the agreement between \texttt{GPT-4o} scores and the aggregated human ratings across evaluation dimensions.  The resulting agreement scores are as follows: Faithfulness (\mbox{$\kappa$ = 0.582}), Relevance ($\kappa$ = 0.643), Informativeness ($\kappa$ = 0.633), Accessibility ($\kappa$ = 0.624), Explanation Usefulness ($\kappa$ = 0.615), and Explanation Accuracy ($\kappa$ = 0.597). We find that human raters and \texttt{GPT-4o} are in substantial agreement. Following this, we expand the evaluation to include all samples in the test sets across our three datasets. We present results on SciNews in Figure~\ref{fig:gpt4o_evaluation_scinews} and in the Appendix for the other two datasets (see figures~\ref{fig:gpt4o_evaluation_elife} and~\ref{fig:gpt4o_evaluation_plos}).

\begin{figure}[t]
  \centering
  \includegraphics[width=0.45\textwidth]{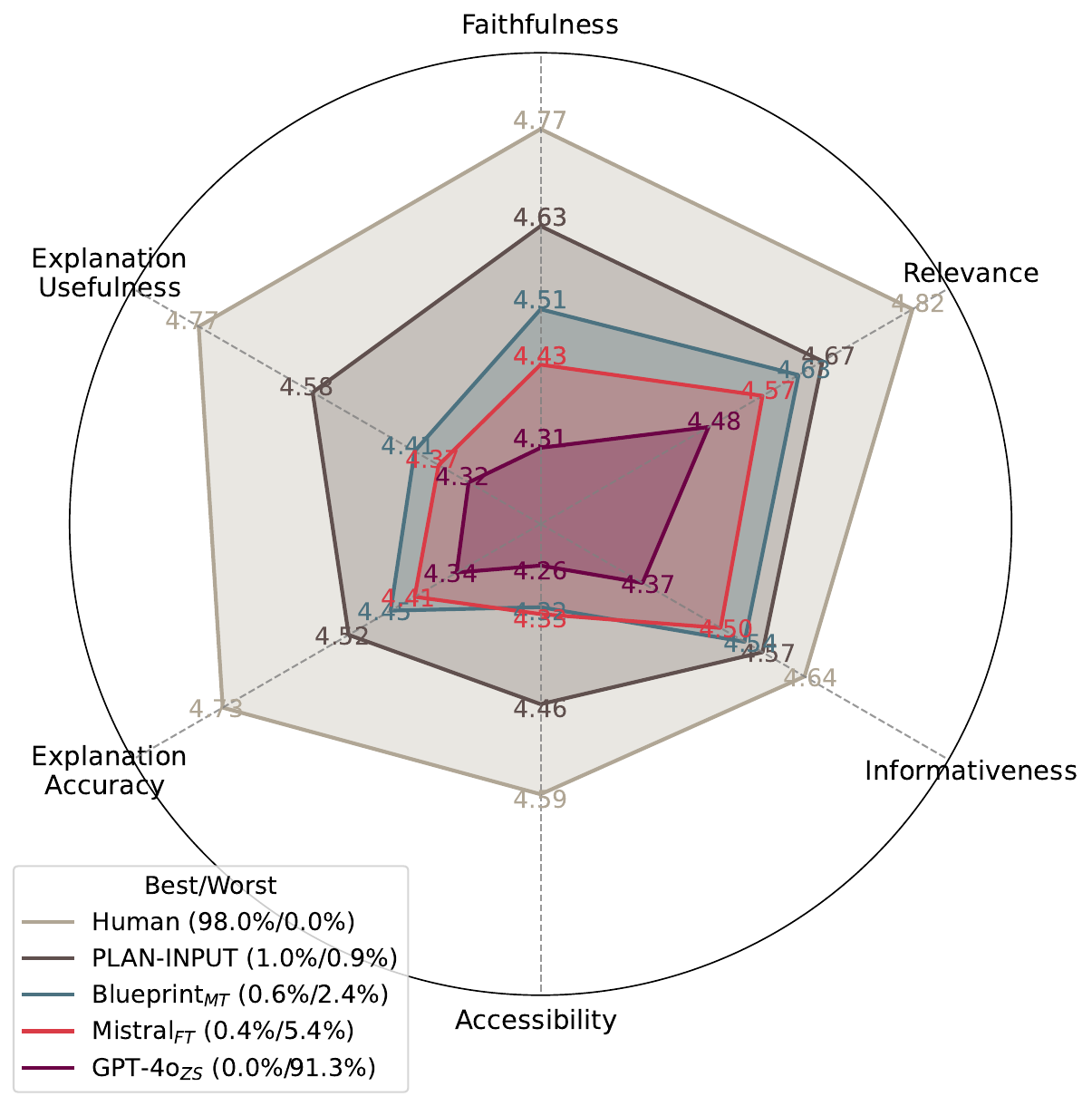}
  \caption{Evaluation with \texttt{GPT-4o} on SciNews.}
  \label{fig:gpt4o_evaluation_scinews}
\end{figure}

\texttt{GPT-4o} assigns the lowest scores to its own answers compared to fine-tuned models. Human-written summaries consistently receive the highest rating and are generally regarded as of the highest quality. In line with our human evaluation, \mbox{\texttt{GPT-4o}} also recognizes that \texttt{Plan-Input} is better than \texttt{Blueprint}$_{MT}$ and \texttt{Mistral}$_{FT}$. We discuss the statistical significance of our results in Appendix~\ref{significance_test}.

\section{Qualitative Analysis}

We conduct a qualitative analysis in Appendix \ref{case_study} by examining two case studies (see Tables \ref{tab:case1} and \ref{tab:case2}), with a particular focus on comparing outputs generated by humans and machines. Through this analysis, we observe differences in hallucination patterns across different models, and our findings suggest that incorporating structured planning into the generation process can effectively reduce certain types of errors.

Specifically, plan-based models demonstrate improved factual consistency compared to \texttt{GPT-4o} and \texttt{Mistral}$_{FT}$ by better organizing information before generation. For instance, when describing how the cerebellum adjusts movements based on feedback (see Table~\ref{tab:case1}), \texttt{GPT-4o} states, ``The cerebellum acts as the brain's command center for motor control.'' At first glance, this phrasing may seem reasonable, but it inaccurately represents the cerebellum's role and could lead to misconceptions. Instead of commanding movements, the cerebellum fine-tunes them using feedback from previous actions, adjusting their timing and coordination. When summarizing a document focusing on Musashi-1, a protein that binds to molecules of RNA and helps to promote cell growth during development (see Table~\ref{tab:case2}), \texttt{Mistral}$_{FT}$ incorrectly attributes certain properties to Musashi-1, such as it being ``permanently destroyed Musashi-1'' by oleic acid. In reality, Musashi-1 is not permanently destroyed but rather inhibited or down-regulated by compounds like oleic acid. These mistakes suggest a tendency to overgeneralize patterns or infer causal relations that are not supported by evidence. In contrast, plan-based models (\texttt{Plan-Output} and \texttt{Plan-Input}) avoid these types of errors.

Unlike \texttt{GPT-4o} and \texttt{Mistral}$_{FT}$, which often produce fragmented or overly general statements, our models demonstrate a better grasp of causal and logical connections within the scientific narrative. For example, in the cerebellum case (see Table~\ref{tab:case1}), our models successfully preserve the logical flow between experimental observations and their implications, while \texttt{GPT-4o} distorts causal relations by overgeneralizing.  This is because structured planning forces the model to first outline key arguments and causal relations before generating text. This process reduces the risk of introducing spurious causal or conceptual connections, which are particularly important for explanation-oriented text. 

Plan-based models are not without limitations; for instance, we occasionally observe misattributions, such as linking Musashi-1 to kidney cancer (see Table~\ref{tab:case1}). However, they offer an improvement over models like GPT-4o in both factual accuracy and overall reliability. In sum, the qualitative analysis underscores the potential of plan-based models to enhance lay summarization quality while also highlighting areas for further refinement.

%% file: sections/conclusion.tex
\section{Conclusion}
In this paper, we propose a planning-driven method designed for lay summarization. Our approach leverages insights from Rhetorical Structure Theory and the Question Under Discussion framework to create question-based plans tailored to explanatory content. We develop two models that differ in how they integrate planning in the summarization process (plans are part of the source document or the target summary) and investigate their impact on overall performance. Empirical studies demonstrate that our models improve summarization performance compared to vanilla fine-tuning and surpass previous state-of-the-art methods. Furthermore, unlike phrase-based planning models, our analysis confirms the effectiveness of utilizing discourse knowledge for high-level planning, resulting in more factually consistent and higher-quality summaries.

\section*{Acknowledgements}
This project has received funding from the European Research Council (ERC) under the European Union’s Horizon 2020 Research and Innovation Programme (Grant Agreement No. 948878). We acknowledge the inspiring environment of TRR 248 funded by DFG (German Research Foundation) – Project Number 389792660. Lapata acknowledges the support of the UK Engineering and Physical Sciences Research Council (Grant EP/W002876/1). We are grateful to the anonymous reviewers and action editor (Junyi Jessy Li) for their exceptionally detailed and helpful feedback.

\begin{figure}[h] 
\centering
\includegraphics[width=0.75\columnwidth]{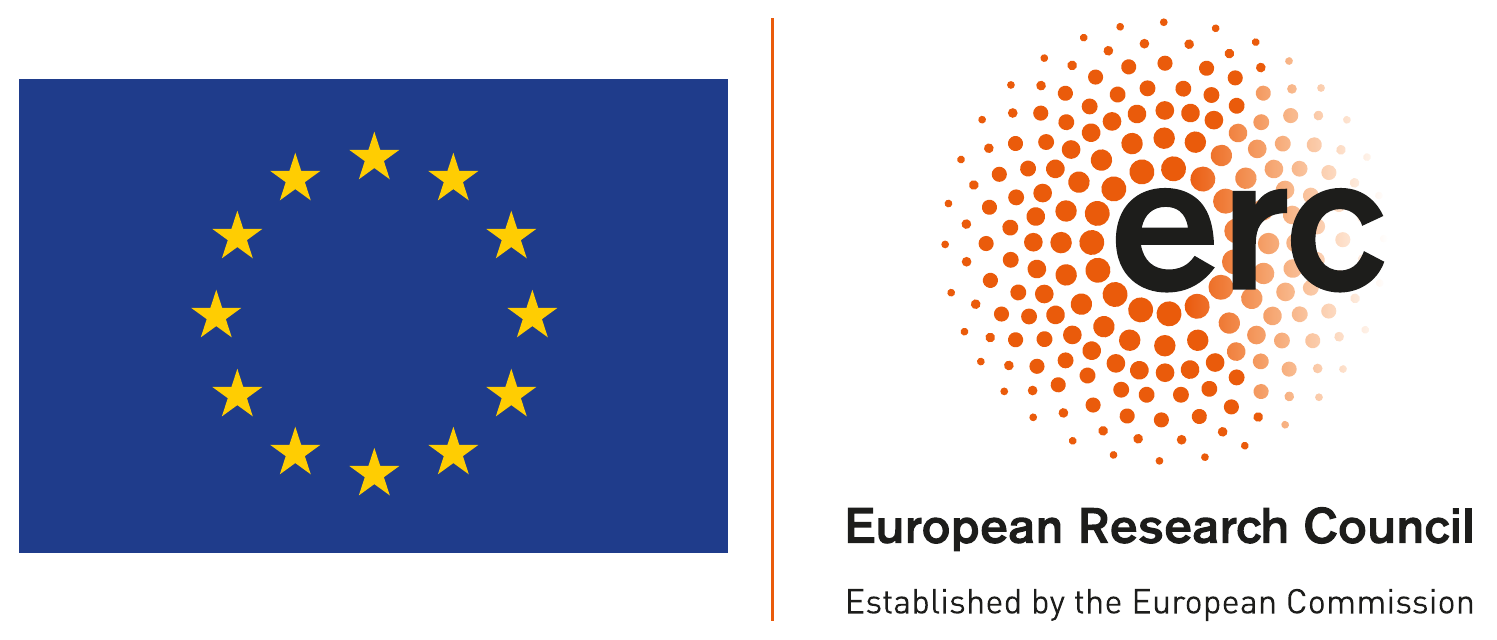}
\end{figure}

%% file: sections/appendix.tex
\section{RST Labels}
\label{app:rst_labels}

\begin{table}[h]
\centering
\begin{tabular}{cP{4.5cm}}
\toprule
\textsc{RST type} & \multicolumn{1}{c}{\textsc{RST label}} \\
\midrule
\textit{Background} & Background, Circumstance \\
\textit{Elaboration} & Elaboration-additional, Elaboration-general-specific, Elaboration-part-whole, Elaboration-process-step, Elaboration-object-attribute, Elaboration-set-member, Example, Definition \\
\textit{Explanation} & Evidence, Explanation-argumentative, Reason \\
\textit{Comparison} & Comparison, Preference, Analogy, Proportion and Topic-Comment \\
\bottomrule
\end{tabular}
\caption{RST explanatory relations.}
\label{tab:RST_explanatory_label}
\end{table}

We will now explain in detail the semantics of the discourse relations listed in Table~\ref{tab:RST_explanatory_label}, and how they can serve as explanations. For instance, the \textit{Background} relation is used to set the context for an event, such as ``The storm caused massive flooding. As a result, the government declared a state of emergency.'' Here, the background information helps to explain why a state of emergency is declared. Similarly, the \textit{Elaboration} relation provides additional information to clarify a statement, as in ``The new software version is more user-friendly. It includes a redesigned interface and improved performance,'' where the additional details elaborate on why the software is considered user-friendly. Other elaboration types include general-specific, part-whole, process-step, object-attribute, set-member, and example, each providing specific details to support and explain a broader statement. For example, ``Many animals hibernate during winter. For example, bears often find caves or dig dens to sleep through the cold month'' elaborates on a general statement with a specific example.

\textit{Explanation} relations, such as \textit{Evidence} and \textit{Reason}, offer support or justification for a statement. For instance, ``Many species are endangered due to habitat destruction. Studies have shown that deforestation has led to significant population decline.'' uses evidence to support the claim about endangered species. Similarly, ``She missed the meeting because her car broke down'' provides a reason to explain her absence.  \textit{Comparison} relations like \textit{Preference} and \textit{Analogy}, help explain by highlighting similarities or differences, as in ``A computer's memory works like a human brain, storing information for later use,'' where the analogy clarifies the concept of computer memory by comparing it to a familiar idea.

\verbdef{\regex}{differ(?:s|ed|ing)}
\begin{table*}[t]
\centering
\footnotesize
\begin{tabularx}{\textwidth}{>{\raggedright\arraybackslash}p{3cm}X}
\hline
\textbf{Category} & \textbf{Signal Words and Phrases} \\
\hline
\textbf{Background} & \textasciicircum historically, \textasciicircum traditionally, \textasciicircum previously, \textasciicircum in the past, \textasciicircum before, \textasciicircum initially, \textasciicircum once, \textasciicircum earlier, \textasciicircum in the beginning, \textasciicircum at first, \textasciicircum prior to, \textasciicircum originally, \textasciicircum at the outset, \textasciicircum at the time, \textasciicircum long ago, \textasciicircum decades ago, \textasciicircum in former times, \textasciicircum previously mentioned, \textasciicircum the history of, \textasciicircum the origin of, \textasciicircum in earlier times, \textasciicircum from the outset, \textasciicircum in the early days, \textasciicircum over the years, \textasciicircum long before, \textasciicircum centuries ago, \textasciicircum during the early stages, \textasciicircum at that time, \textasciicircum back then, \textasciicircum once upon a time, \textasciicircum throughout history, \textasciicircum previously established, \textasciicircum over the course of history, \textasciicircum in ancient times \\
\hline
\textbf{Comparison} & compared to, compared with, \textasciicircum similarly, likewise, in contrast, in comparison, in opposition, \textasciicircum on the contrary, \textasciicircum on one hand, \textasciicircum on the other hand, \textasciicircum conversely, rather than, different from, \textasciicircum unlike, similar to, analogous to, contrary to, in contradistinction, distinct from, distinguishable from, as opposed to, in the same way, by comparison, comparable to, \regex from, diverg(e|es|ed|ing)? from, in a similar manner, in the same vein, on the flip side, correspondingly, on a different note, in opposition to, different from, in a contrasting way, in an analogous way \\
\hline
\textbf{Elaboration} & defined as, refer(?:s|red|ring)? to, mean(?:s|t)?, known as, definition, \textasciicircum in other words, \textasciicircum that is to say, \textasciicircum that's to say, \textasciicircum this is to say, \textasciicircum that means, \textasciicircum this means, \textasciicircum this implies, \textasciicircum that implies, i.e., e.g., for example, for instance, such as, \textasciicircum to clarify, \textasciicircum to explain, whereas, \textasciicircum to illustrate, \textasciicircum to elaborate, \textasciicircum specifically, \textasciicircum particularly, in particular, as an example, by way of example, more precisely, \textasciicircum to be specific, \textasciicircum to exemplify, namely, by way of illustration, expounded upon, in more detail, one example, an example, \textasciicircum to add to this \\
\hline
\textbf{Explanation} & because, due to, since, thanks to, owing to, for the sake of, stemming from, given that, in light of, for this reason, for that reason, for the reason that, for the purpose of, for this cause, the reason is, the reasons are, as a result, consequently, as a consequence, accordingly, with the result that, so that, such that, result(?:s|ed|ing)? in, , result(?:s|ed|ing)? from, lead(?:s|ed|ing)? to, which means, thereby, whereby, in consequence of, on account of, so as to, on the grounds that \\
\hline
\end{tabularx}
\caption{Words and phrases used to heuristically identify discourse relations signaling explanations.}
\label{tab:signal_words}
\end{table*}

\section{Hyper-parameter Settings}
\label{hyperparameter_settings}
For all fine-tuning experiments, we utilize the AdamW optimizer \cite{loshchilovdecoupled} with $\beta_1$ = 0.9, $\beta_2$ = 0.999, $\epsilon$ = $10^{-9}$, and weight decay = 0.1, along with a warm-up ratio of 0.15. The initial learning rate is set to 5e-5, employing cosine learning rate scheduling. We set the random seed to~2,024 and the dropout rate to~0.1. For  SciNews, we set the maximum number of new generation tokens to 1,024, for eLife to~512, and for PLOS to~256.  All other parameters follow the default settings of the \texttt{Transformers} library. During training, we save the checkpoint with the highest Rouge-2~F1  on the validation set as the final checkpoint. All experiments run for 15~epochs with a batch size of 8, and early stopping is implemented to prevent overfitting, with all models converging before reaching 15 epochs. During inference, we employ beam search with  size~4,  length penalty of~3.0, and set the no-repeat n-gram size to~3.

For \texttt{GPT-4o},\footnote{\url{https://platform.openai.com/docs/models/}} we use version \textit{gpt-4o-2024-05-13}  (our experiments took place May 1, 2024 to March 31, 2025) with the following hyperparameter settings: We set temperature to~1, top\_p to~1, frequency penalty to~0.2, and presence penalty to~0.2. All other hyperparameters follow OpenAI's default values.

\section{Rule-based Extraction of Explanations}
\label{rule_based_method}
We develop a set of rules aimed at identifying words and phrases that indicate four types of explanations, which are in turn based on RST relations:

\begin{itemize}[leftmargin=8pt,itemsep=1pt,topsep=1pt,parsep=1pt]
    \item \textbf{Background}  includes \emph{Background} and \emph{Circumstance} relations, signalled by words such as ``historically'' and ``traditionally.''
    \item \textbf{Elaboration} covers various \emph{Elaboration} relations (e.g.,~\emph{Additional}, \emph{General-Specific}, \emph{Part-Whole}, \emph{Process-Step}, \emph{Object-Attribute}, and \emph{Set-Member}), as well as \emph{Example} and \emph{Definition} relations, signalled by words like ``defined as'' and ``for example.''
    \item \textbf{Explanation} includes \emph{Evidence}, \emph{Explanation-Argumentative}, and \emph{Reason} relations, signalled by words such as ``because'' and ``due to.''
    \item \textbf{Comparison} includes \emph{Comparison}, \emph{Preference}, \emph{Analogy}, and \emph{Proportion} relations, signalled by words like ``compared to'' and ``similarly.''
\end{itemize}

Table \ref{tab:signal_words} presents the words and phrases associated with each of these relations. To identify these indicators within the text, we compile regular expression patterns tailored to each category. Explanatory content is collected on a sentence-by-sentence basis: If a signal word is detected within a sentence, that entire sentence is categorized accordingly. In constructing these patterns, we use the caret symbol \verb|^| before certain words to restrict their occurrence to sentence-initial positions. This method ensures that words like ``historically'' or ``similarly'' are recognized only when they function as introductory elements, which is common in explanations. Additionally, we utilize regular expressions such as \texttt{differ(?|ed|ing)} to capture different word forms. Once each explanatory sentence is identified, we then select the preceding sentence as its target sentence.

\section{Model Prompts}
\label{app:prompts}
In this section, we present the various prompts used in our work.

\begin{figure}[htpb]
\centering
\begin{tcolorbox}[colback=gray!10!white,colframe=black!50!black,title=Question Generation Prompt,fonttitle=\bfseries, halign title=flush center, width=0.5\textwidth]
Generate a question that is coherent and contextually relevant to the provided context and target sentence. \\
Context: \{\texttt{Context Text}\}\\
Target: \{\texttt{Target Sentence}\} \\
Question Sentence:
\end{tcolorbox}
\caption{Prompt used by \texttt{GPT-4o} and \texttt{Mistral} to generate plan questions based on target sentences and their context.}
\label{GPT-4o/mistral_Generate_Question_from_Answer}
\end{figure}

\begin{figure}[htpb]
\centering
\begin{tcolorbox}[colback=gray!10!white,colframe=black!50!black,title=Summary Generation Prompt, fonttitle=\bfseries, halign title=flush center, width=0.5\textwidth]
Generate a lay summary for the provided document. \\
Document: \{\texttt{Document Text}\} \\
Lay Summary:
\end{tcolorbox}
\caption{Prompt used by \texttt{Mistral}$_{ZS}$, \texttt{GPT-4o}$_{ZS}$ and \texttt{Mistral}$_{FT}$ to generate summaries.}
\label{prompt_zero_shot_learning}
\end{figure}

\begin{figure}[htpb]
\centering
\begin{tcolorbox}[colback=gray!10!white,colframe=black!50!black,title=Summary Generation Prompt (ICL),fonttitle=\bfseries, halign title=flush center,  width=0.5\textwidth]
Document: \{\texttt{Document Text}\}\\
Lay Summary: \{\texttt{Summary Text}\}\\
Using the example above as a reference for structure and tone, generate a new lay summary for the following document. Ensure the summary is original and does not replicate phrases or content from the example. \\
Document: \{\texttt{Document Text}\} \\
Lay Summary:
\end{tcolorbox}
\caption{Prompt used by \texttt{Mistral}$_{ICL}$ and \texttt{GPT-4o}$_{ICL}$ to generate summaries.}
\label{prompt_in_context_learning}
\end{figure}

\begin{figure}[htpb]
\centering
\begin{tcolorbox}[colback=gray!10!white,colframe=black!50!black,title=Named Entity Identification,fonttitle=\bfseries, halign title=flush center, width=0.5\textwidth]
{Identify all named entities in the following text and return them as a list.\\
Document: \{\texttt{Document Text}\} \\
Return Format: \texttt{[Entity 1, 2, 3]} \\
List all identified entities in the order they appear in the text.\\
Identified Entities:}
\end{tcolorbox}
\caption{Prompt used by \texttt{GPT-4o} to identify named entities for \texttt{Blueprint} model.}
\label{prompt_named_entities}
\end{figure}

\begin{figure}[htpb]
\centering
\begin{tcolorbox}[colback=gray!10!white,colframe=black!50!black,title=Plan-based Summary Generation Prompt,fonttitle=\bfseries, halign title=flush center, width=0.5\textwidth]
Generate a lay summary for the following document based on the plan questions.\\
Document: \{\texttt{Document Text}\} \\
Planning Questions: \{\texttt{Questions}\} \\
Ensure that the generated summary sequentially answers the plan questions.\\
Lay Summary:
\end{tcolorbox}
\caption{Prompt used by \texttt{Plan-Input} and \texttt{Plan-Output} models.}
\label{prompt_fine_plan_tuning_learning}
\end{figure}

\begin{figure}[t]
\centering
\begin{tcolorbox}[colback=gray!10!white,colframe=black!50!black,title=Direct Extraction of Explanatory Content,fonttitle=\bfseries, halign title=flush center, width=0.5\textwidth]
Parse the provided document to extract all explanatory sentences and the corresponding target sentences they explain. Return the parsing results as a list of dictionaries formatted as follows:

\begin{verbatim}
[
  {
    explanatory_sentence: 'XXX',
    target_sentence: 'XXX'
  },
  {
    explanatory_sentence: 'XXX',
    target_sentence: 'XXX'
  }
]
\end{verbatim}
Ensure that each dictionary accurately pairs each explanatory statement with its respective target sentence.\\
Document: \{\texttt{Document Text}\}
\end{tcolorbox}
\caption{Prompt used by \texttt{GPT-4o} and \texttt{Mistral} to directly extract explanatory content.}
\label{GPT-4o/mistral_as_RST_Parser}
\end{figure}

\begin{figure}[htpb]
\centering
\begin{tcolorbox}[colback=gray!10!white,colframe=black!50!black,title=Irrelevant Question Generation,fonttitle=\bfseries, halign title=flush center, width=0.5\textwidth]

Randomly generate a question with a question mark. \\ 
Question Sentence: 
\end{tcolorbox}
\caption{Prompt used by \texttt{GPT-4o} to generate irrelevant questions.}
\label{Irrelevant_Question_Generation}
\end{figure}

\onecolumn
\section{Human Evaluation Guidelines}
\label{human_evaluation_guideline}
\begin{figure*}[h]
\footnotesize
\centering
\begin{tcolorbox}[]

\paragraph{Prerequisites} Eligibility for this evaluation requires simultaneous fulfillment of two conditions: (1) being a Master's or Ph.D. student in Computer Science or Computational Linguistics, and (2) demonstrating greater than or equal to C2 English proficiency\footnote{\url{https://en.wikipedia.org/wiki/C2_Proficiency}}. If you do not meet both criteria, we respectfully ask you to refrain from participating in this task. Those who qualify are encouraged to proceed and follow the instructions below.

\paragraph{Instructions} Below is a detailed explanation of the metrics and evaluation criteria for our human evaluation process. Please carefully read the provided document along with the candidate summaries. After thoroughly examining each summary, evaluate them based on the following six criteria using a Likert scale from 1 to 5, with higher scores reflecting better quality:

\begin{itemize}[leftmargin=8pt,itemsep=1pt,topsep=1pt,parsep=1pt]
    \item \textbf{Faithfulness:} This metric evaluates how accurately the summary reflects the information in the source document. A faithful summary should strictly adhere to the source material, avoiding any contradictions or unverified details.
    \item \textbf{Relevance:} This metric assesses how accurately the summary content reflects the topics covered in the source text. A relevant summary should include topics that are pertinent to the source document.
    \item \textbf{Informativeness:} This metric measures how well the summary conveys the key points and essential details from the source text. An informative summary should capture the main ideas, providing a clear and precise understanding of the source document's arguments and findings.
    \item \textbf{Accessibility:} This metric evaluates how easy the summary is to read and understand. An accessible summary should be well-structured and written in clear language. It should avoid unnecessary complexity and ensure that readers can follow the content without difficulty.
    \item \textbf{Explanation Accuracy:} This metric measures how factually correct the explanatory content is. Explanation accuracy requires that the information provided in the explanatory sentence is verifiable and aligns with established knowledge in the relevant field, avoiding any misleading or incorrect explanations.
    \item \textbf{Explanation Usefulness:} This metric assesses how helpful the explanatory content is in enhancing the reader's understanding of the subject matter. Explanation usefulness ensures that the content contributes to the reader's comprehension of complex ideas, offering valuable insights and clarifications.
\end{itemize} 
Please note that you are required to highlight all explanatory content found in the summary before assigning evaluation scores. You are permitted to use external knowledge sources, such as books, academic papers, and Wikipedia, but you are not allowed to use any AI tools to assist in your judgment. If you encounter any specific explanatory content that you are unable to evaluate for accuracy or usefulness, you should skip that content and mark it (meaning it will not factor into your scoring, but you will still need to score based on the rest of the content). This will allow us to bring in external experts (PhD holders in the relevant field) to assess it.

\paragraph{Rating System}
For each criterion, use the Likert scale as follows:
\begin{itemize}[leftmargin=8pt,itemsep=1pt,topsep=1pt,parsep=1pt]
    \item 1 (Worst): Very poor quality, does not meet the criteria at all.
    \item 2 (Poor): Subpar quality, meets the criteria to a minimal extent.
    \item 3 (Fair): Average quality, adequately meets the criteria.
    \item 4 (Good): Above average quality, meets the criteria well.
    \item 5 (Best): Excellent quality, fully meets the criteria.
\end{itemize}
\paragraph{Overall Ranking}
After rating the summaries based on the six criteria, you are also expected to rank the candidates from best to worst based on overall quality. Consider how well each summary performs across all criteria to determine the final ranking.
\end{tcolorbox}
\caption{A snapshot of the experimental instructions seen by our human raters. }
\end{figure*}

\twocolumn
\section{Human Expert Evaluation Results}
\label{Expert_Results}
Our human evaluators identify six examples where they could not make a decision on the accuracy and usefulness of explanations. Three are generated by \texttt{GPT-4o}$_{ZS}$, two by \texttt{Mistral}$_{FT}$, and one by \texttt{Blueprint}$_{MT}$, covering fields such as Biology, Neuroscience, and Materials Science. To further assess these cases, we enlist experts with a Ph.D. degree in the relevant areas. Using the same guidelines, \texttt{GPT-4o} receives an average rating of 2.33 for explanation accuracy and 2.67 for explanation usefulness. For \texttt{Mistral}$_{FT}$, the average rating is~3.50 for both accuracy and usefulness, while \texttt{Blueprint}$_{MT}$ receives a rating of~4.00 for accuracy and 2.00 for usefulness. These expert evaluations align with the initial human assessment, reinforcing the consistency of the findings across the highlighted explanations.

\section{\texttt{GPT-4o} Evaluation Results}

\begin{figure}[h]
  \centering
  \includegraphics[width=0.42\textwidth]{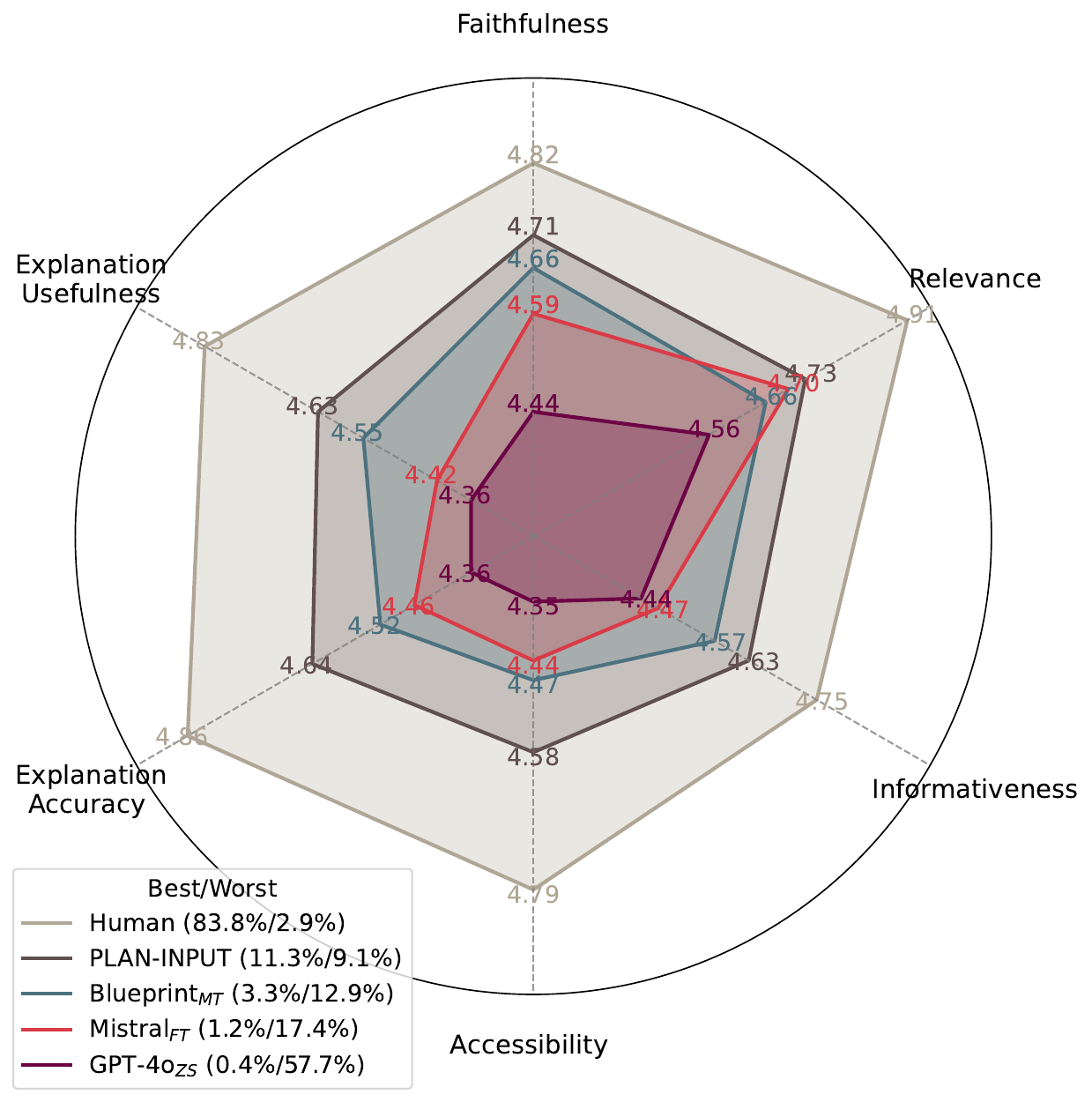}
  \caption{Evaluation results (dimensions of Faithfulness, Relevance, Informativeness, Accessibility, Explanation Accuracy, and Explanation Usefulness) on eLife using \mbox{\texttt{GPT-4o}} as a judge.}
  \label{fig:gpt4o_evaluation_elife}
\end{figure}

\begin{figure}[ht!]
  \centering
  \includegraphics[width=0.42\textwidth]{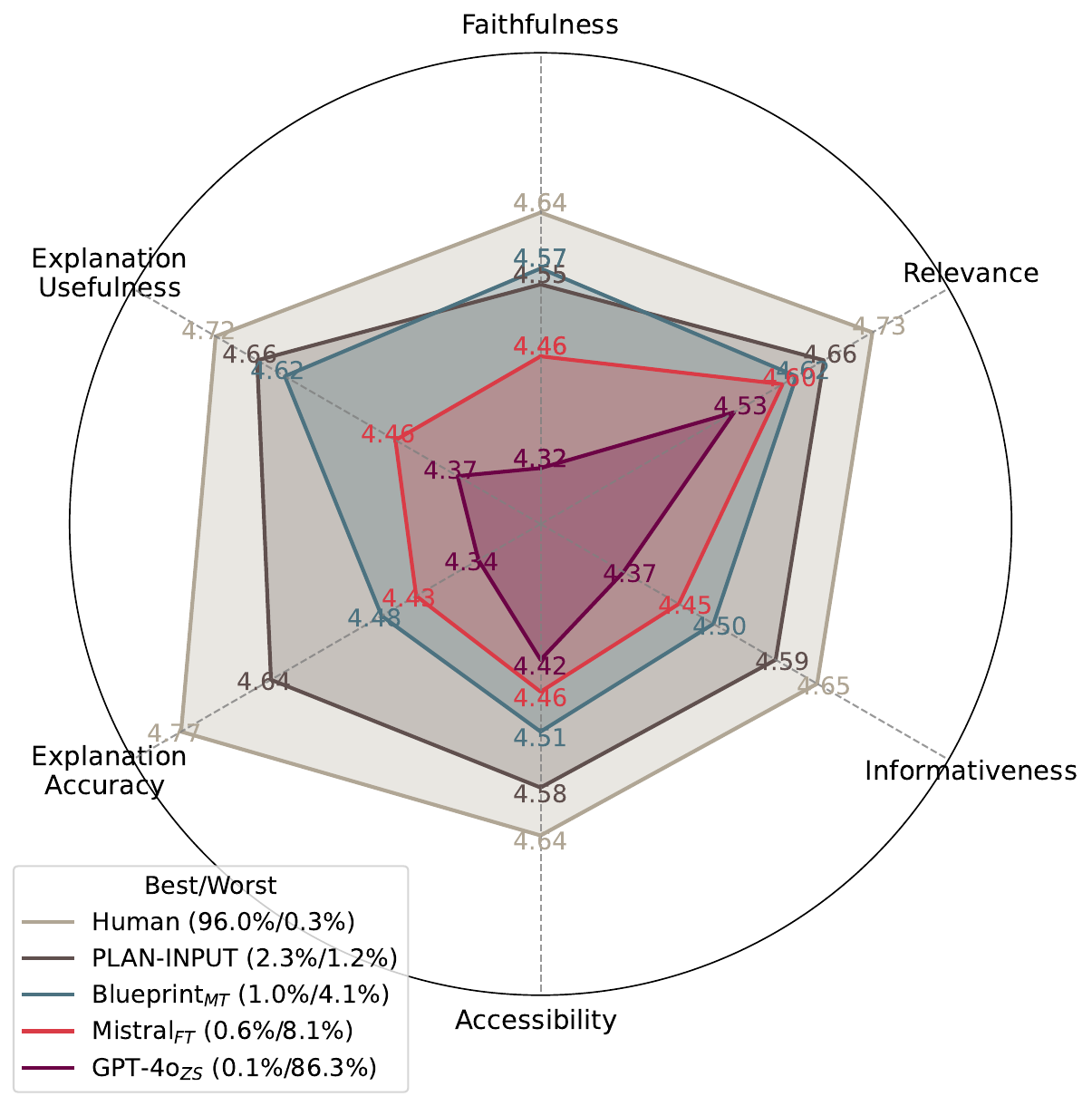}
  \caption{Evaluation results (dimensions of Faithfulness, Relevance, Informativeness, Accessibility, Explanation Accuracy, and Explanation Usefulness) on PLOS using \mbox{\texttt{GPT-4o}} as a judge.}
  \label{fig:gpt4o_evaluation_plos}
\end{figure}

Figure~\ref{fig:gpt4o_evaluation_elife} shows automatic evaluation results on the eLife dataset using \texttt{GPT-4o} as the judge. Results on the PLOS dataset are summarized in Figure~\ref{fig:gpt4o_evaluation_plos}, again using \texttt{GPT-4o} as the evaluator. 

\section{Significance Testing}
\label{significance_test}
We use paired t-tests to analyze the results of our human evaluation and automatic evaluation (LLM-as-Judge with \mbox{\texttt{GPT-4o}}). In the human evaluation, we find that human responses are perceived as significantly better against all baseline models across metrics;  with regard to  Faithfulness and Accessibility,  differences against \texttt{Plan-Input} and \texttt{Blueprint}$_{MT}$  are not statistically significant. \texttt{Plan-Input} is significantly better in Explanation Usefulness and Accessibility compared to all neural network models (but not human answers).

In the LLM-as-Judge evaluation, human summaries outperform all neural models across all datasets and metrics with statistical significance. \texttt{Plan-Input} is significantly better over other neural network models on the SciNews dataset across all metrics except Relevance and Informativeness. On the eLife dataset, \texttt{Plan-Input} outperforms other neural network models on all metrics except Faithfulness, Explanation Usefulness, and Relevance. On the PLOS dataset, \texttt{Plan-Input} significantly outperforms competitor models in all metrics except Informativeness, Faithfulness, and Relevance. These results are derived with a significance level of \( p < 0.05 \).

\section{Case Study}
\label{case_study}
In this case study, we compare human-written lay summaries with those generated by neural models (see Tables \ref{tab:case1} and \ref{tab:case2}). Human summaries go beyond simply listing experimental results; they provide well-structured interpretations of the research background, methods, and key findings, whereas neural models often generate content that appears \textit{plausible but contains inaccuracies}. 

\begin{table*}[hbp]
\centering
\scalebox{1}{\tabcolsep=3pt
\begin{threeparttable}
\begin{tabular}{|p{3cm}|p{12cm}|}
\midrule
\scriptsize Model & \scriptsize Output \\
\midrule
\scriptsize Human Reference & \scriptsize The cerebellum utilizes proprioceptive feedback to fine-tune the timing of movements in a sequence based on previous actions. Imagine the cerebellum as a coach who watches how you perform a move, then gives tips to improve the next one based on what was seen. But how exactly does it achieve this? To investigate, we trained rabbits to blink in response to an external cue and explored whether the cerebellum could use feedback from one blink to trigger the next. As expected, after learning the initial blink, the rabbits blinked again in response to their own first blink, creating a chain of movements. Control experiments confirmed that each blink was initiated by the previous one rather than the original cue. Consistent patterns of brain activity during this process indicate that the cerebellum adjusts movement based on feedback from previous actions. Building on this, we trained rabbits to blink on cue, and they learned to initiate additional blinks in response to earlier blinks in the sequence. We further found that the rabbits could use a blink from one eye as a cue to trigger a blink in the other eye, suggesting that the same mechanism governs these movements. This raises the possibility that the cerebellum might also guide sequences of cortical activity during cognitive tasks, given its extensive connections to the cortex, a question future experiments should explore.\\
\hdashline

\scriptsize \texttt{GPT-4o} & \scriptsize 
The cerebellum acts as the brain’s \textcolor{red}{command center for motor control}, directly issuing movement instructions and ensuring precise timing. To test this, scientists trained rabbits to blink in response to a \textcolor{red}{vibration signal}. Once trained, the rabbits began blinking in a continuous rhythm, even without an external cue, indicating that the cerebellum could independently generate movement sequences. Interestingly, when the researchers covered one of the rabbit’s eyes, the uncovered eye still continued blinking, suggesting that the cerebellum maintains a direct, hardwired link between the two eyes. \textcolor{red}{These findings demonstrate that the cerebellum is not only responsible for movement execution but may also function as a memory hub for motor programs, completely independent of the cerebral cortex.}\\
\hdashline

\scriptsize \texttt{Mistral$_{FT}$} & \scriptsize The cerebellum helps coordinate movements by adjusting actions based on prior attempts. To study this, scientists trained rabbits to blink in response to a sound. Over time, the rabbits developed a habit of blinking twice in succession, \textcolor{red}{suggesting that the cerebellum is responsible for generating repetitive movements}. Further experiments showed that rabbits could use one blink as a cue for another, even switching between eyes. This suggests that the cerebellum not only controls basic motor sequences but also plays a role in \textcolor{red}{synchronizing different body movements}. It is even possible that the cerebellum helps regulate \textcolor{red}{general cognitive functions}, though more research is needed. \\
\hdashline

\scriptsize \texttt{Plan-Output} & \scriptsize 

q1: What can the role of the cerebellum in motor control be compared to? \newline q2: How do rabbits form a sequence of consecutive blinks after training?  \newline q3: What does it suggest that rabbits can use a blink from one eye to trigger a blink in the other eye?

\rule[0.5ex]{\linewidth}{0.4pt}

The cerebellum acts like a coach, refining movements based on \textcolor{red}{past experiences}. To explore this, scientists trained rabbits to blink when they heard a tone. Over time, the rabbits learned to initiate a second blink in response to their own first blink, forming a sequence of movements. This suggests that the cerebellum uses sensory feedback to fine-tune motor control. Furthermore, the rabbits could blink one eye and then trigger a blink in the other eye, indicating that the cerebellum directly controls \textcolor{red}{bilateral facial coordination}. These findings highlight the cerebellum’s essential role in sequencing motor actions and suggest that it may also contribute to higher-level cognitive processes. \\
\hdashline
\scriptsize \texttt{Plan-Input} & \scriptsize 

q1: What kind of role does the cerebellum play in motor control?\newline q2: How do rabbits trigger a sequence of blinks from an initial blink after training? \newline q3: What does it reveal about motor control mechanisms that rabbits can use a blink from one eye to trigger a blink in the other eye?

\rule[0.5ex]{\linewidth}{0.4pt}

The cerebellum functions as a predictive system that learns movement patterns and automatically sequences them. Researchers trained rabbits to blink when given an air puff. After sufficient training, the rabbits began to produce consecutive blinks, with each blink triggering the next. This suggests that the cerebellum can \textcolor{red}{store movement chains and replay them when needed}. Additionally, the researchers found that a blink in one eye could lead to a blink in the opposite eye, implying that the cerebellum might have specialized \textcolor{red}{circuits} dedicated to synchronizing both sides of the body. These insights open the possibility that the cerebellum also orchestrates sequences of thoughts and decisions, a topic for future research. \\
\midrule
\end{tabular}
\end{threeparttable}
}
\caption{Case study 1: We use red font to mark hallucinated information. This sample comes from the test split of the eLife dataset \cite{goldsack-etal-2022-making}.}
\label{tab:case1}
\end{table*}

\begin{table*}[hbp]
\centering
\scalebox{1}{\tabcolsep=3pt
\begin{threeparttable}
\begin{tabular}{|p{3cm}|p{12cm}|}
\midrule
\scriptsize Model & \scriptsize Output \\
\midrule
\scriptsize Human Reference & \scriptsize When an embryo is developing, stem cells must divide and develop into many specialized types of cells. However, if cell division doesn't stop, or if it restarts later in life, it can cause tumors to form. Musashi-1 is a protein that binds to molecules of RNA and helps to promote cell growth during development: mice that lack this protein have serious brain defects and die shortly after birth. Musashi-1 is usually turned off in adult cells that are not dividing. Sometimes, however, it remains active and contributes to the growth of cancers in the brain and the gut. Reducing Musashi-1 levels in colon tumors slows their growth and causes the cancer cells to die. To find a compound that would switch off Musashi-1, Clingman et al. screened more than 30,000 compounds and identified four inhibitors. One of these was oleic acid, a fatty acid that is found in olive oil and other animal and plant oils. Oleic acid interacts with Musashi-1 in a way that changes the shape of the protein. These changes mean that Musashi-1 is no longer able to regulate the genes that control cell proliferation. Clingman et al. also found that Musashi-1 promotes the activity of a particular enzyme that makes fatty acids; molecules that are needed in large quantities when cells are dividing. Musashi-1 appears to act as a nutrient sensor, turning down the activity of this enzyme in cells when levels of oleic acid are high, and turning up enzyme activity when oleic acid levels are low. The findings of Clingman et al. further reveal how our diets can affect gene expression, and may aid the development of new treatments against cancer.\\
\hdashline

\scriptsize \texttt{GPT-4o} & \scriptsize Musashi-1 is a protein that plays a crucial role in early development by directly binding to \textcolor{red}{DNA} and activating genes responsible for cell proliferation. In mice that lack Musashi-1, the protein is compensated by other molecular pathways, allowing them to develop \textcolor{red}{normally} with only minor cognitive impairments. To explore ways to inhibit Musashi-1, Clingman et al. analyzed a vast library of nearly \textcolor{red}{135,000} chemical compounds. They identified four potential inhibitors, including oleic acid, a fatty acid mainly found in \textcolor{red}{red meat and dairy products}. This compound works by \textcolor{red}{permanently destroying} Musashi-1, preventing it from affecting cell division. In cancer studies, researchers found that reducing Musashi-1 levels in colon tumors leads to \textcolor{red}{uncontrolled tumor growth and increased metastasis}. These findings suggest that Musashi-1 may function as a \textcolor{red}{critical tumor suppressor}, and future therapies should aim to \textcolor{red}{enhance its activity in cancer patients rather than inhibit it.} \\
\hdashline

\scriptsize \texttt{Mistral$_{FT}$} & \scriptsize Musashi-1 is a protein that regulates cell division by binding to RNA. It is active during embryonic development but usually turns off in adults. However, it can remain active and contribute to tumors in the brain and \textcolor{red}{lungs}. To inhibit Musashi-1, researchers screened over 30,000 compounds and found oleic acid, a substance present in \textcolor{red}{coconut oil and dairy products}. Oleic acid interacts with Musashi-1 and \textcolor{red}{permanently destroys it}, stopping it from regulating genes linked to cell growth. Additionally, high oleic acid levels \textcolor{red}{completely shut down metabolism in cells}, showing how diet influences gene regulation and cancer treatment. \\

\hdashline

\scriptsize \texttt{Plan-Output} & \scriptsize 

q1: What is the role of Musashi-1 in cell development and cancer growth? \newline q2: How does oleic acid affect Musashi-1 and its function in cell proliferation? \newline q3: How does Musashi-1 function as a nutrient sensor in relation to fatty acid synthesis?

\rule[0.5ex]{\linewidth}{0.4pt}

Musashi-1 is a protein that binds to RNA molecules and plays a crucial role in cell division and tissue development. During early embryonic growth, it ensures proper differentiation of stem cells. Mice lacking Musashi-1 experience severe neurological defects and typically survive for \textcolor{red}{several months} before succumbing to developmental issues. Normally, Musashi-1 is turned off in mature cells, but in some cases, it remains active and has been linked to tumor formation in the brain and \textcolor{red}{kidneys}. To counteract Musashi-1 activity, Clingman et al. screened over 30,000 potential inhibitors and identified four promising compounds. One of them, oleic acid—found in olive oil and fish oil—was particularly effective. It interacts with Musashi-1 in a way that causes the protein to \textcolor{red}{break down}, thereby preventing it from controlling genes involved in cell proliferation. Additionally, Musashi-1 functions as a metabolic sensor by regulating fatty acid synthesis. When oleic acid levels are low, Musashi-1 suppresses enzyme activity, reducing fatty acid production. However, when oleic acid concentrations rise, the enzyme becomes highly active, increasing fatty acid availability to support rapid cell growth. These findings illustrate the link between diet and gene regulation and could inform new therapeutic strategies for cancer treatment. \\
\hdashline
\scriptsize \texttt{Plan-Input} & \scriptsize 

q1: What happens to mice that do not have Musashi-1, and what does this indicate about its function? \newline q2:  How did researchers identify inhibitors of Musashi-1, and what is one of the inhibitors they found? \newline q3: What effect does reducing Musashi-1 levels have on colon tumors, and what does this suggest for cancer treatment?

\rule[0.5ex]{\linewidth}{0.4pt}

Musashi-1 is a crucial protein that binds to RNA and plays a key role in cell division and development. When mice lack Musashi-1, they develop mild brain defects but survive into \textcolor{red}{adulthood}, though with noticeable neurological impairments. This suggests that Musashi-1 is essential for normal brain development and proper neural function. To identify potential inhibitors of Musashi-1, Clingman et al. conducted a large-scale screening of nearly \textcolor{red}{52,000} chemical compounds. Among the four inhibitors they discovered, one was oleic acid, a fatty acid commonly found in olive oil and coconut oil. Oleic acid interacts with Musashi-1 by binding to it directly, leading to its complete deactivation. Reducing Musashi-1 levels in colon tumors has been shown to accelerate their growth, \textcolor{red}{making the tumors more aggressive}.  This suggests that Musashi-1 may have a dual role in cell regulation, potentially acting as both a tumor promoter and suppressor in different contexts. These findings provide new insights into the complex role of Musashi-1 in cancer and could inform the development of novel therapeutic strategies.  \\
\midrule
\end{tabular}
\end{threeparttable}
}
\caption{Case study 2: We use red font to mark hallucinated information. This sample comes from the test split of the eLife dataset \cite{goldsack-etal-2022-making}.}
\label{tab:case2}
\end{table*}